\documentclass[times, 10pt]{elsarticle}

\usepackage{hyperref}

\usepackage{multirow}
\usepackage{longtable}
\usepackage{booktabs}
\usepackage{subfig}
\usepackage{mathrsfs} 
\usepackage{tikz}
\usepackage{graphicx}
\usepackage{forest}
\usetikzlibrary{shadows}
\usepackage{amsmath,amsfonts}
\usepackage{algorithmic}
\usepackage{algorithm}
\usepackage{array}
\usepackage{url}
\usepackage{pifont}
\usepackage{verbatim}
\usepackage{multirow}
\usepackage{booktabs}
\usepackage{bm}
\usepackage{makecell}
\usepackage{color} 
\usepackage{adjustbox}
\usepackage{rotating}
\usepackage{tabularx}
\PassOptionsToPackage{rgb}{xcolor}
\usetikzlibrary{mindmap}
\journal{Pattern Recognition}
\begin{document}
	\tikzset{
	my node/.style={
		draw=gray,
		inner color=gray!5,
		outer color=gray!10,
		thick,
		minimum width=1cm,
		% rounded corners=3,
		text height=1.5ex,
		text depth=0ex,
		font=\sffamily,
		drop shadow,
	}
}

\begin{frontmatter}

\title{Domain-Shared Learning and Gradual Alignment for Unsupervised Domain Adaptation Visible-Infrared Person Re-Identification}

\author[1,2]{Nianchang Huang}
\ead{huangnianchang@xidian.edu.cn}
\author[1,2]{Yi Xu}
\ead{yi\_xu@stu.xidian.edu.cn}
\author[1,2]{Ruida Xi}
\ead{ruidaxi@stu.xidian.edu.cn}
\author[1,2]{Qiang Zhang*}
\ead{qzhang@xidian.edu.cn}

\address[1]{State Key Laboratory of Electromechanical Integrated Manufacturing of High-Performance Electronic Equipments, Xidian University, Xi’an, Shaanxi 710071, China}

\address[2]{Center for Complex Systems, School of Mechano-Electronic Engineering, Xidian University, Xi’an, Shaanxi 710071, China}

\cortext[cor1]{Corresponding author:Qiang Zhang.}
%\fntext[equal]{These authors contributed equally to this work.}

\begin{abstract}
Recently, Visible-Infrared person Re-Identification (VI-ReID) has achieved remarkable performance on public datasets. However, due to the discrepancies between public datasets and real-world data, most existing VI-ReID algorithms struggle in real-life applications. To address this, we take the initiative to investigate Unsupervised Domain Adaptation Visible-Infrared person Re-Identification (UDA-VI-ReID), aiming to transfer the knowledge learned from the public data to real-world data without compromising accuracy and requiring the annotation of new samples. Specifically, we first analyze two basic challenges in UDA-VI-ReID, \emph{i.e.}, inter-domain modality discrepancies and intra-domain modality discrepancies. Then, we design a novel two-stage model, \emph{i.e.}, Domain-Shared Learning and Gradual Alignment (DSLGA), to handle these discrepancies. In the first pre-training stage, DSLGA introduces a Domain-Shared Learning Strategy (DSLS) to mitigate ineffective pre-training caused by inter-domain modality discrepancies via exploiting shared information between the source and target domains. While, in the second fine-tuning stage, DSLGA designs a Gradual Alignment Strategy (GAS) to handle the cross-modality alignment challenges between visible and infrared data caused by the large intra-domain modality discrepancies through a cluster-to-holistic alignment way. Finally, a new UDA-VI-ReID testing method \emph{i.e.}, CMDA-XD, is constructed for training and testing different UDA-VI-ReID models. A large amount of experiments demonstrate that our method significantly outperforms existing domain adaptation methods for VI-ReID and even some supervised methods under various settings.

\end{abstract}


\end{frontmatter}

\section{Introduction}
%Person Re-Identification(ReID) algorithms\cite{Ye_Shen_Lin_Xiang_Shao_Hoi_2022}  aims to identify the given pedestrians across different scenes, which facilitates extensive applications in camera surveillance and public safety. Therefore, ReID has garnered widespread research interest and is rapidly evolving. Specifically, in the early stage, ReID mainly focuses on identifying pedestrians from visible images \emph{i.e.}, VV-ReID. Although VV-ReID has achieved good performance, visible cameras fail to capture discriminative information under low-light conditions, thus easily leading to VV-ReID failure. In contrast, infrared cameras can effectively obtain enough information under such low-light conditions. As a result, infrared cameras are increasingly being integrated into modern surveillance systems, making Visible-Infrared Person Re-Identification (VI-ReID) \cite{Wu_Zheng_Yu_Gong_Lai_2017} a mainstream approach in recent years. This advancement has significantly expanded the applicability of ReID algorithms.

Recently, VI-ReID algorithms have obtained significant improvements, which have surpassed humans by a large margin on some public datasets \cite{Wu_Zheng_Yu_Gong_Lai_2017, Nguyen_Hong_Kim_Park_2017}. However, directly applying those VI-ReID algorithms to real-life applications usually results in a significant decline in accuracy because of the huge distribution discrepancies between training data from publicly available datasets and real-world ones. A straightforward way to address such an issue is to label images in real-life applications and re-train those VI-ReID algorithms. However, the cost of annotating a large dataset is usually very high. Naturally, a question is raised: \emph{can we develop strategies to transfer the knowledge learned from public data to real-world data without compromising accuracy and requiring new annotations?} Unsupervised Domain Adaptation ReID (UDA-ReID) is such a task that aims to transfer the learned knowledge from the source domain to the target domain. However, existing UDA-ReID methods\cite{Zheng_Lan_Zeng_Zhang_Zha_2020,
Li_Zhang_2020} mainly focuses on single-modality data, \emph{i.e.}, UDA-VV-ReID. Unsupervised Domain Adaptation VI-ReID (UDA-VI-ReID) models are rarely investigated in existing research. Considering that, we take the initiative to investigate UDA-VI-ReID in this paper. 
\begin{figure}[t]
\centering
\includegraphics[width=3in]{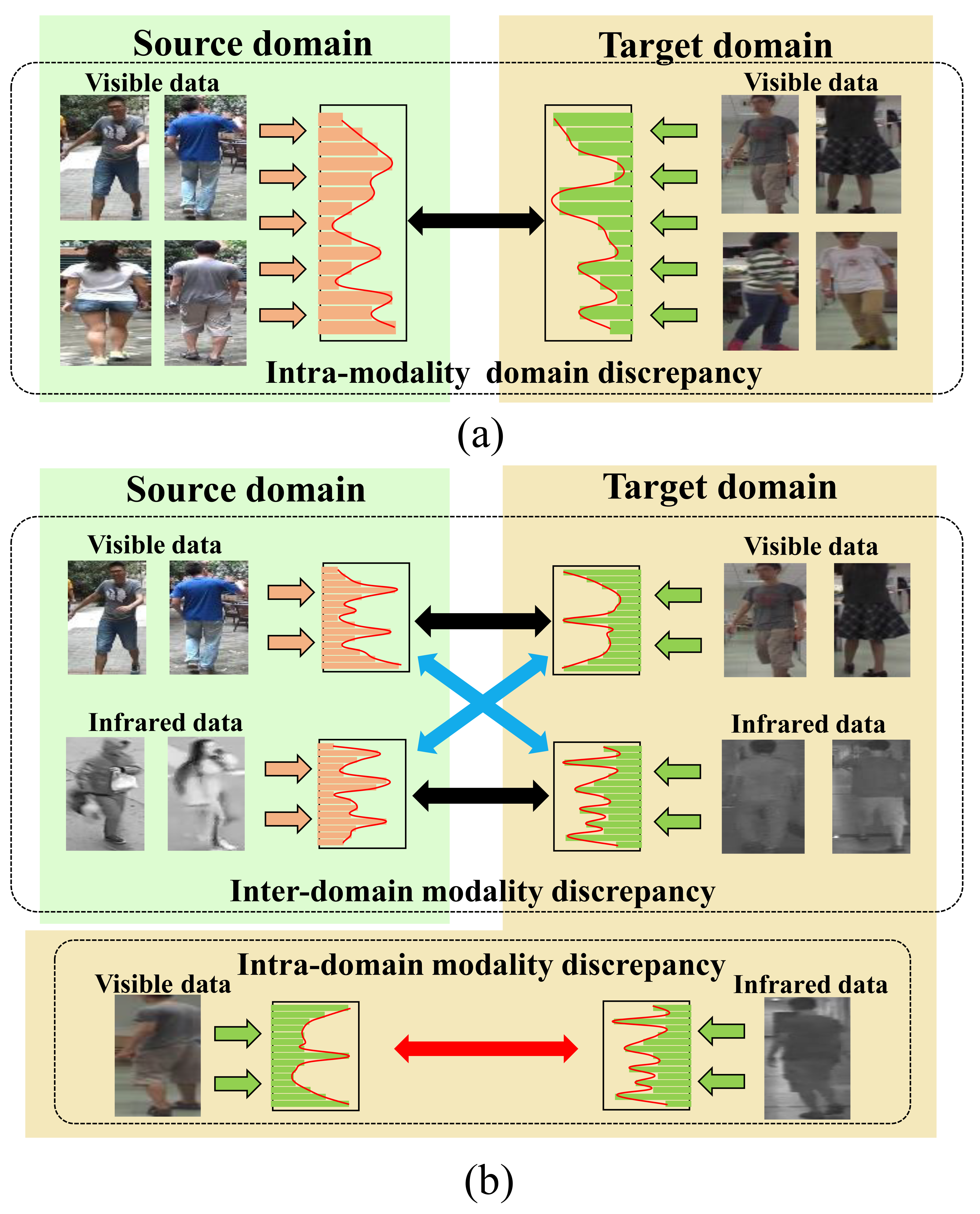}
\caption{
Comparisons between UDA-ReID and UDA-VI-ReID. (a) The intra-modality domain discrepancies in UDA-ReID. (b) The inter-domain modality discrepancies and intra-domain modality discrepancies in UDA-VI-ReID. Arrows in different colors mean different types of discrepancies
%The black arrows denote the domain discrepancies across the source and target domain within the same modality, the blue arrows denote the domain discrepancies across the source and target domains among different modalities, and the red arrow denotes the modality discrepancies within the same domain.
}
\label{fig_discrepancy}
\end{figure}

As shown in Fig. \ref{fig_discrepancy} (b), the inter-domain modality discrepancies are the major challenge in UDA-VI-ReID, which refers to that UDA-VI-ReID simultaneously encounters intra-modality and inter-modality domain discrepancies across the source and target domains (\emph{i.e.}, the black and blue arrows, respectively, in Fig. \ref{fig_discrepancy}(b)), rather than just intra-modality domain discrepancies as in single-modality UDA-ReID (\emph{i.e.}, the black arrow in Fig. \ref{fig_discrepancy} (a)).
Meanwhile, the intra-domain modality discrepancies refer to the fact that even within the same domain in UDA-VI-ReID, the modality discrepancies will also hinder the learning of those discriminative person features (\emph{i.e.}, the red arrow in Fig. \ref{fig_discrepancy} (b)), especially when the ground-truth labels of the target domain data are absent during training.

The inter- and intra-domain modality discrepancies will make existing single-modality UDA-ReID methods \cite{tian2018cr, Zheng_Lan_Zeng_Zhang_Zha_2020,
Li_Zhang_2020} suffer in UDA-VI-ReID if without any modifications. To address these challenges, this paper proposes a preliminary solution for UDA-VI-ReID based on the classic clustering-based UDA-ReID framework, 
%Specifically, the clustering-based framework is one of the mainstream solutions for single-modality UDA-ReID and has achieved great progress
which usually first pre-trains a ReID network on the source domain data via some supervised learning strategies, and then fine-tunes the pre-trained network on the target domain data, following a clustering-based pseudo label generation and pseudo label-based network optimization iterative paradigm.

However, such a framework faces the following issues when directly applied to the UDA-VI-ReID task. First, the VI-ReID network pre-trained on the source domain data may suffer a decline in feature extraction capability when used on target domain data due to inter-domain modality discrepancies. Especially, the feature extraction ability of a VI-ReID network pre-trained on the source domain data may be only slightly better than that of a randomly initialized VI-ReID network, when directly applied to the target domain data. Secondly, the core of the fine-tuning stage is to generate high-quality pseudo labels for unlabeled target domain data and use them to optimize the performance of the pre-trained network. While generating pseudo labels within the same modality is relatively straightforward, aligning labels of the same identity across modalities (\emph{i.e.}, addressing the cross-modality alignment challenges) is much more difficult due to significant intra-domain modality discrepancies. Those incorrectly matched visible-infrared clusters will easily lead to training bias and sub-optimal results.  
To address the aforementioned issues, this paper proposes a novel two-stage UDA-VI-ReID method, namely the Domain Shared Learning and Gradual Alignment (DSLGA). 

In the pre-training stage, DSLGA realizes that, beyond their own specific information, the source and target domains inherently share some common information (\emph{e.g.}, person shapes and contours), and accordingly designs a Domain-Shared Learning Strategy (DSLS) to leverage such domain-shared information for reducing their inter-domain modality discrepancies. Specifically, DSLS first projects the features extracted from the two domains, respectively, into the same feature space to capture their domain-shared information by using a parameter-shared VI-ReID network. Then, DSLS specially designs a Domain-Shared Adversarial Loss (DSAL) to pull close the distributions of the source and target domains in a united feature space via a generative adversarial way, thus facilitating the extraction of domain-shared information. Moreover, DSLGA incorporates a Cluster Refinement with Multiple Results (CRMR) module, which is specifically designed to generate more reliable intra-modality pseudo labels for each modality within the target domain. 

In the fine-tuning stage, DSLGA introduces a Gradual Alignment Strategy (GAS) to gradually overcome intra-domain modality discrepancies, spanning from cluster-level alignment for cross-modality pseudo label generation to holistic-level alignment for VI-ReID network optimization. Specifically, GAS first presents a Supplementary Graph Matching (SGM) module to tackle cross-modality alignment challenges and generate high-quality cross-modality pseudo labels for target domain data. Especially, beyond the classic paradigm inter-modality alignment, SGM innovatively proposes an intra-modality supplementary alignment step for those cross-modality unaligned clusters, aiming to realign them with their closest intra-modality clusters.
This further exploits such unaligned clusters, especially for those being over-clustered ones. Then, when training the VI-ReID network, GAS further designs a Cross-Modality Consistency Constraining (CMCC) module to suppress those incorrect pseudo labels that are inevitably produced by SGM. The core of CMCC lies in utilizing all person information within the source and target domains (\emph{i.e.}, holistic information) to assess the confidence levels of the generated pseudo labels, subsequently suppressing those pseudo labels with lower confidence and reinforcing those with higher confidence.

Furthermore, we also construct a new CMDA-XD testing method on top of some existing cross-modality datasets for UDA-VI-ReID in this paper. Overall, the contributions of this paper can be summarized as follows:
\begin{itemize}
\item[$\bullet$] This paper takes the initiative to investigate the unsupervised domain adaptation VI-ReID, aiming to transfer the knowledge learned from the source domain data to the target domain data without requiring the annotation of new samples. Accordingly, a novel UDA-VI-ReID model DSLGA is proposed and a new testing method CMDA-XD is constructed in this paper.
\item[$\bullet$] We design a DSLS in the pre-training stage to alleviate the inter-domain modality discrepancies and obtain a well-initialized network for the target domain by exploring the domain-shared information between the source and target domains.
\item[$\bullet$] We present a GAS to overcome the large intra-domain modality discrepancies through a cluster-to-holistic alignment way, thus significantly boosting the performance of our model on target domain data.
\end{itemize}

\section{Related Work}
\subsection{Supervised Visible-Infrade Person Re-Identification}
Most existing VI-ReID models can be divided into two categories: modality-specific information compensation based models \cite{zhang2022fmcnet} and modality-shared feature learning based models \cite{ye2021channel, ye2018visible}. 
Specifically, modality-specific information compensation based models follow a two-stage strategy. They first handle cross-modality discrepancies by generating information for the missing modalities from the existing ones, thus enabling them to achieve (RGB-T)-to-(RGB-T) matching. Then, they capture discriminative cross-modality complementary information from paired RGB-T images to address intra-modality variations. For example, Zhang  \emph{et al.} \cite{zhang2022fmcnet} proposed feature-level modality-specific information generation rather than image-level generation for VI-ReID. For that, they first decompose the single-modality RGB (IR) features into modality-specific RGB (IR) features and modality-shared RGB (IR) features. Then, they generate the missing modality-specific IR (RGB) features from those decomposed modality-shared  RGB (IR) features by presenting a Feature-level Modality Compensation (FMC) module. Eventually, they combine three types of features for matching. 

Differently, modality-shared feature learning based models try to simultaneously address the intra-modality variations and inter-modality variations by capturing those discriminative person-related features co-existing in the two modalities. For example, Liu \emph{et al}. \cite{liu2021sfanet} developed a shareable dual-path
information-preserving network to extract 3D-shaped middle-level shareable feature vectors, aiding the model in learning discriminative feature representations.

However, the current VI-ReID algorithms are tailored to specific datasets. When these algorithms are applied to real-world applications, their accuracies will experience a notable decline due to the substantial distribution variations between public data and real-world data. In this paper, we will introduce a pioneering UDA-VI-ReID task to address this issue.

\subsection{Unsupervised Domain Adaptation Person Re-Identification}
Generally speaking, existing UDA-ReID models can be divided into two categories: domain transformation-based models \cite{tian2018cr} and pseudo label-based models \cite{Zheng_Lan_Zeng_Zhang_Zha_2020,
Li_Zhang_2020}. Domain transformation-based models are based on the idea of image style transformation, which tries to reduce domain discrepancies by migrating the image styles from the source domain to the target domain. During this process, their learned knowledge in the source domain will be also transferred to the target domain, thus being able to rightly identify different persons for the target domain data.
For example, Huang  \emph{et al.} \cite{Huang_Wu_Xu_Zhong_Zhang_2021} used a Suppression of Background Shift Generative Adversarial Network (SBSGAN) to generate images with the background being suppressed, thus mitigating the data gaps between the source and target domains. They further proposed a Densely Associated 2-Stream (DA-2S) network with an update strategy to learn discriminative ID features from the generated data, which considers both the human-body information and certain useful ID-related cues in the background.

On the other hand, pseudo label-based models \cite{
Li_Zhang_2020} first pre-train a model with source domain data via supervised methods, and then iterate them between clustering and fine-tuning on the target domain data. For instance, Chen \emph{et al.}\cite{Yixiao_Chen_Li_2020} designed a Mutual Mean-Teaching (MMT) based on a pseudo label generation strategy, which uses clustering algorithms to assign labels to every sample and the teacher-student model to suppress data noise. 
He \emph{et al.} \cite{He_Shen_Guo_Ding_Guo_2022} focused on the pseudo label generation strategy during the fine-tuning stage, since the qualities of those generated pseudo labels determine fine-tuning results. They argued that the consistency among different feature spaces is the key to the pseudo label's qualities. Accordingly, a Self-Consistent pseudo label Refinement method (SECRET) was proposed to improve the consistency by mutually refining the pseudo labels generated from different feature spaces, resulting in more accurate pseudo labels.

Although existing UDA-ReID models have achieved great
progress, those models will not work when facing the large
inter- and intra-domain modality discrepancies within UDA-VI-ReID. Considering that,this paper takes
the initiative to investigate Unsupervised Domain Adaption
Visible-Infrared person Re-Identification (UDA-VI-ReID).

\subsection{Unsupervised Domain Adaptation Visible-Infrared Person Re-Identification}

Recently, some other researchers also studied UDA-VI-ReID. Specifically, 
Yang \emph{et al.} \cite{Yang_Chen_Ma_Ye} proposed a Translation, Association and Augmentation (TAA) framework to transfer learned knowledge from the labeled visible source domain to the unlabelled visible-infrared target domain.
For that, they first employed a modality translator to transfer visible images into infrared images, formulating generated visible-infrared image pairs for cross-modality supervised training. Then, a Robust Association and Mutual Learning (RAML) module and a Translation Supervision and Feature Augmentation (TSFA) module were designed to resist label noise and enhance feature discriminability, respectively. 

Different from TAA that transfers the learned knowledge from the labeled visible source domain to the unlabeled visible-infrared target domain, this paper focuses on transferring the learned knowledge from the labeled visible-infrared source domain to the unlabeled visible-infrared target domain.

\section{Methodology}
\begin{figure*}[!t]
\centering
\includegraphics[width=\linewidth]{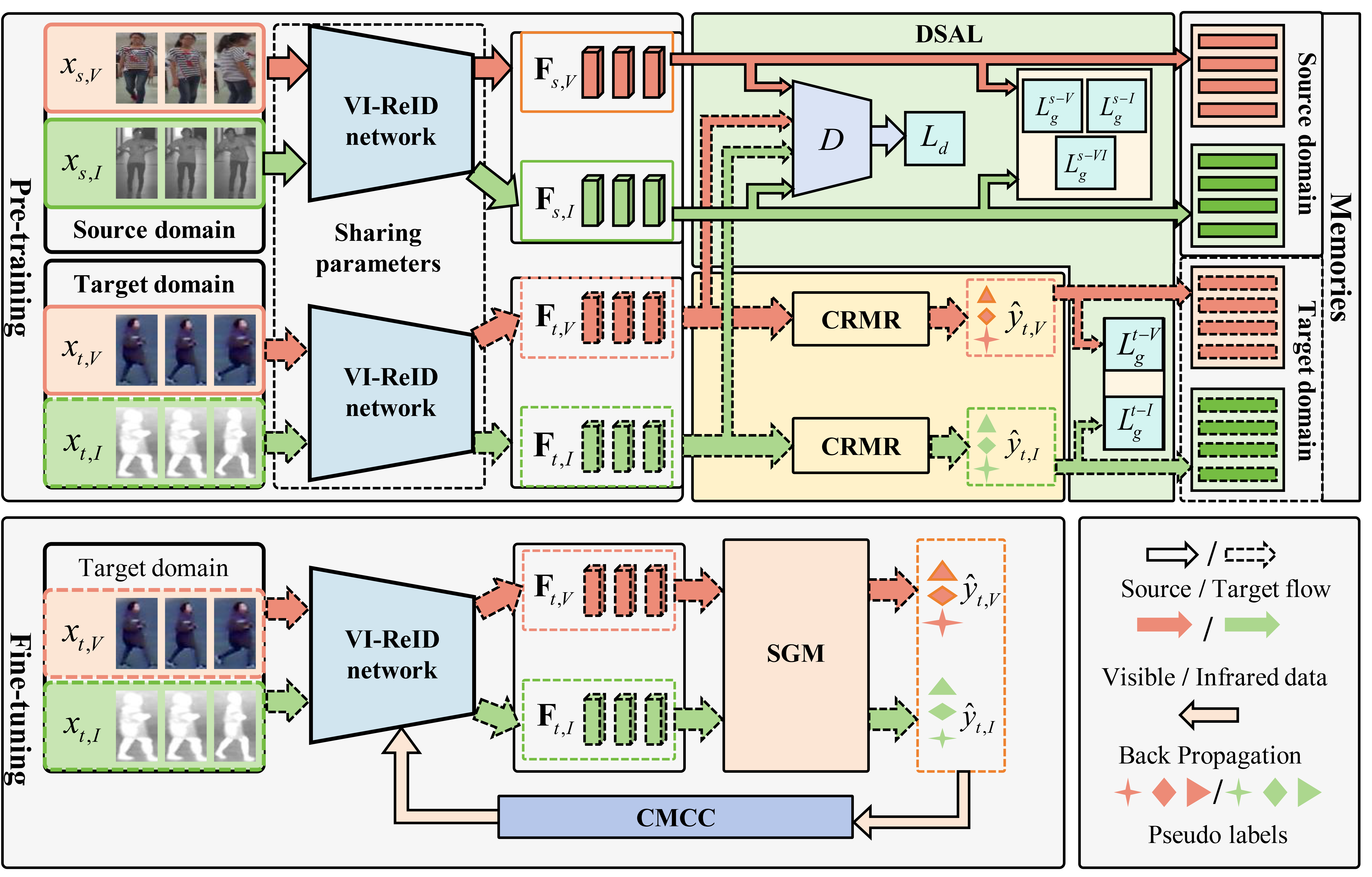}
\caption{Framework of our proposed DSLGA model. 
It contains two main stages, \emph{i.e.}, pre-training and fine-tuning. In the pre-training stage, a VI-ReID network is trained to achieve the knowledge transfer from the source domain to the target domain. Especially, a domain-shared learning strategy (DSLS) is designed to mitigate the inter-domain modality discrepancies with the aid of the proposed DSAL. As well, a CRMR is proposed to generate intra-modality pseudo labels for the target domain. In the fine-tuning stage, the pre-trained model in the first state is further optimized by deeply exploring the target domain data. Especially, a gradual alignment strategy (GAS) is designed to mitigate the intra-domain modality discrepancies, which first generates cross-modality pseudo labels at the cluster level by an SGM module and then suppresses those incorrect pseudo labels at the holistic level by a CMCC module. 
}
\label{framework}
\end{figure*}
The overall framework of our proposed Domain-Shared Learning and Gradual Alignment (DSLGA) is shown in Fig. \ref{framework}, which mainly contains two stages: pre-training and fine-tuning. In the pre-training stage, a Domain-Shared Learning Strategy (DSLS) is designed to explore the domain-shared information across the source and target domains to address the inter-domain modality discrepancies. Especially, DSLS will design a Domain-Shared Adversarial Loss (DSAL) to pull close the distributions
of the source and target domains in a united feature space
via a generative adversarial way. Besides, a Cluster Refinement with Multiple Results (CRMR) module will be specifically designed to generate pseudo labels for addressing the label deficiency problem in the target domain.
In the fine-tuning stage, a Gradual Alignment Strategy (GAS) will be presented to achieve cross-modality alignment, overcoming the intra-domain modality discrepancies within the target domain data. Especially, a Supplementary Graph Matching (SGM) module and a Cross-Modality Consistency Constraining (CMCC) module will be proposed in GAS to achieve cluster-level and holistic-level alignments, respectively. In the following contents, we will demonstrate the details of each module in our proposed method.

In this paper, we denote the source domain data by $D_{s} =\left \{ D_{s, V}, D_{s, I} \right \}$ and the target domain data by $D_{t} =\left \{ D_{t, V}, D_{t, I} \right \}$, respectively. Here, 
\begin{equation}
D_{s, V}=\left \{ (x_{s,V}^{n},y_{s,V}^{n})\mid n=1,2, \cdots, N_{s, V} \right \}
\end{equation}
represents the $n$-th visible image $x_{s,V}^{n}$ and its corresponding person identity $y_{s,V}^{n}$. $D_{s, I}=\left \{ (x_{s,I}^{n},y_{s,I}^{n})\mid n=1,2, \cdots, N_{s, I} \right \}$ represents
the $n$-th infrared image $x_{s,I}^{n}$ and its corresponding person identity
$y_{s,I}^{n}$. 
$N_{s,V}$ and $N_{s,I}$ denote the numbers of visible and infrared
images within the source domain data, respectively. Moreover, we suppose that there are $C_{s}$ identities
within the source domain data. Accordingly, $y_{s,V}^{n}$ and $y_{s,I}^{n} \in [1, C_{s}]$.
Similarly,
$D_{t,V}=\left \{ x_{t,V}^{n}\mid n=1,2, \cdots, N_{t,V} \right \}$ and $D_{t,I}=\left \{ x_{t,I}^{n}\mid n=1,2, \cdots, N_{t,I} \right \}$ represent the visible images and the infrared images in the target domain, respectively. It should be noted that the identity labels $y_{t,V}$ and $y_{t,I}$ for the visible and infrared images of the target domain data are missed. Therefore, in different training stages, we will generate their pseudo labels
$\hat{y}_{t,V}$ and $\hat{y}_{t,I}$ via some algorithms to replace $y_{t,V}$ and $y_{t,I}$, respectively. The 
numbers of identities within the pseudo labels $\hat{y}_{t,V}$ and $\hat{y}_{t,I}$ are
denoted by $C_{t,V}$ and $C_{t,I}$, respectively. Subsequently, for a better introduction in
the following content, we denote the target domain data
with its corresponding pseudo labels by $\hat{D}_{t} = \left \{ \hat{D}_{t,V}, \hat{D}_{t,I} \right \}$, where $\hat{D}_{t,V}= \left \{ (x_{t,V}^{n}, \hat{y}_{t,V}^{n}) \mid n=1,2, \cdots, N_{t,V} \right \}$ and $\hat{D}_{t,I}= \left \{ (x_{t,I}^{n}, \hat{y}_{t,I}^{n}) \mid n=1,2, \cdots, N_{t,I} \right \}$. Here $\hat{y}_{t,V}^{n} \in [1, C_{t,V}]$ and $\hat{y}_{t,I}^{n} \in [1, C_{t,I}]$.

\subsection{VI-ReID Network}
\begin{figure}[!t]
\centering
\includegraphics[width=3in]{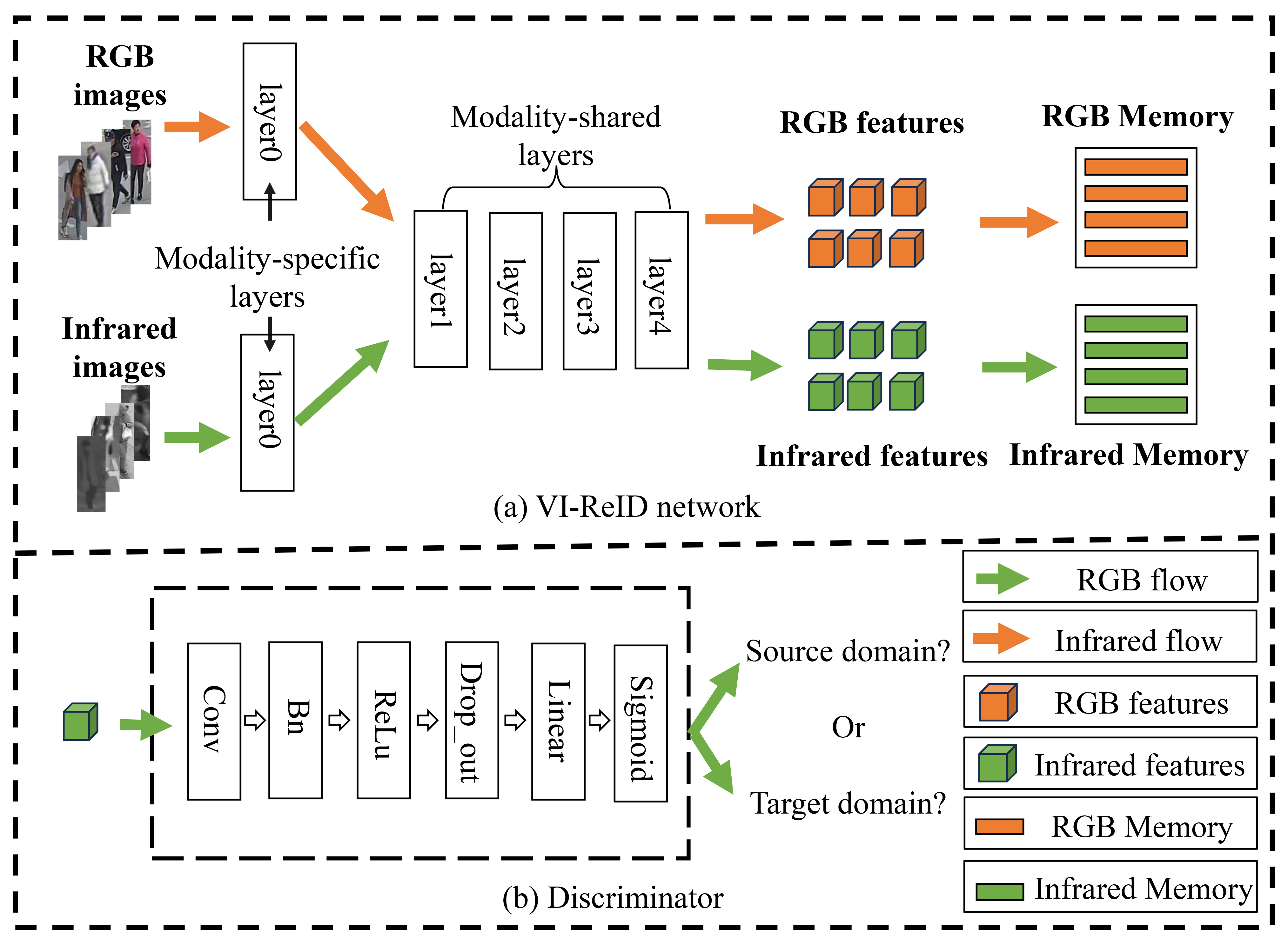}
\caption{
Structures of the VI-ReID network and the discriminator. (a) The VI-ReID network. (b) The discriminator.
Items in different colors represent different person identities.
}
\label{fig_agw}
\end{figure}

As shown in Fig. \ref{fig_agw} (a), our VI-ReID network follows the
classical specific-shared feature extraction structure, which is similar to most existing VI-ReID models\cite{Wu_Zheng_Yu_Gong_Lai_2017, Liu_Sun_Zhu_Pei_Yang_Li}. Specifically, it first employs
two modality-specific feature extraction subnetworks to extract visible features $\mathbf{\hat{F}}_{V}$ and infrared features $\mathbf{\hat{F}}_{I}$ from visible
images and infrared images, respectively, \emph{i.e.},
\begin{equation}
\label{f1}
\mathbf{\hat{F}}_{V} = f_{V,sp}(x_{V}, \phi_{V,sp} ),\mathbf{\hat{F}}_{I} = f_{I,sp}(x_{I}, \phi_{I,sp}),
\end{equation}
where $f_{V,sp}(\ast, \phi_{V,sp})$ and $f_{I,sp}(\ast, \phi_{I,sp})$ denote two modality-specific feature extraction subnetworks with their
parameters $\phi_{V,sp}$ and $\phi_{I,sp}$, respectively. $x_{V}$ and $x_{I}$ denote all visible images and infrared images, respectively, from either the source or target domain data. 
Then, the extracted modality-specific features $\hat{\mathbf{F}}_{V}$ or $\hat{\mathbf{F}}_{I}$ are fed into a modality-shared feature extraction subnetwork to obtain their person features $\mathbf{F}_{V}\in \mathbb{R}^{N_{s,V}\times D}$ ($\mathbf{F}_{V}\in \mathbb{R}^{N_{t,V}\times D}$) or $\mathbf{F}_{I}\in \mathbb{R}^{N_{s,I}\times D}$ ($\mathbf{F}_{I}\in \mathbb{R}^{N_{t,I}\times D}$)
in a modality-shared feature space, \emph{i.e.},
\begin{equation}
%\label{f1}
\mathbf{F}_{V} = f_{sh}( \hat{\mathbf{F}}_{V}, \phi_{sh} ),\mathbf{{\mathbf{F}}}_{I} = f_{sh}(\hat{\mathbf{F}}_{I}, \phi_{sh}).
\end{equation}
Here, $f_{sh}( \ast, \phi_{sh} )$ denotes the modality-shared subnetwork
with its parameters $\phi_{sh}$, which remains unchanged across different modalities. $D$ denotes the channels. For simplicity, we denote the VI-ReID network by $f_{\phi}( \ast, \phi)$ with
$\phi=\left \{ \phi_{V,sp},\phi_{I,sp},\phi_{sh} \right \}$.

Besides the VI-ReID network, two memory modules are employed for the source domain data, and another two memory modules are employed for the target domain data to store the visible and infrared feature centers of each identity, respectively. These stored feature centers
will be employed to assist in different training stages in the
subsequent sections. Considering that the four memory
modules follow the same structure, we take the memory
module constructed for the visible samples in the source domain as an example for introduction in the following contents.

Specially, given the visible images in the source domain
$D_{s, V}$, the memory module
includes a center memory $\mathbb{M}_{s,V} \in \mathbb{R}^{C_{s} \times D}$ along with an initialization operation and an update operation. Here, each
item $\mathbb{M}_{s,V}[c_{s}] \in \mathbb{R}^{1 \times D}$ in  $\mathbb{M}_{s,V}$ denotes
the visible feature center of the $c_s$-th identity in the source
domain data. In different training steps or stages, the center
memory $\mathbb{M}_{s,V}$ is initialized or updated by the following
steps. 

First, the visible features $\mathbf{F}_{s,V}^{{c_s},n_{c_s}} \in \mathbb{R}^{1 \times D}$ for the $n_{c_s}$-th visible image of the $c_s$-th identity are extracted by
\begin{equation}
\mathbf{F}_{s,V}^{{c_s},n_{c_s}}=f_{\phi}( x_{s,V}^{{c_s},n_{c_s}}, \phi),
\end{equation}
where $n_{c_s} \in [1, N_{s, V}^{c_s}]$  and $N_{s, V}^{c_s}$ denotes the total number
of visible images of the $c_s$-th identity. Then, the feature center
$\mathbf{F}_{s,V,center}^{c_s}$ of the $c_s$-th identity is computed by
\begin{equation}
\mathbf{F}_{s,V,center}^{c_s}=\frac{1}{N_{s,V}^{c_s}}\sum_{n_{c_s}=1}^{N_{s,V}^{c_s}} \mathbf{F}_{s,V}^{{c_s},n_{c_s}}.
\end{equation}
For the initialization operation, the center memory
$\mathbb{M}_{s,V}[c_{s}]$ is initialized by
\begin{equation}
\mathbb{M}_{s,V}[c_{s}]=\mathbf{F}_{s,V,center}^{c_s}.
\label{f5}
\end{equation}
While, for the update operation, the center memory $\mathbb{M}_{s,V}[c_{s}]$
is updated by
\begin{equation}
\label{fup}
\mathbb{M}_{s,V}[c_{s}]=\alpha \mathbb{M}_{s,V}[c_{s}]+(1-\alpha)\mathbf{F}_{s,V,center}^{c_s},
\end{equation}
where $\alpha$ denotes the update rate and is set to 0.5 in this paper.

Similarly, we can obtain the infrared feature center memory with $C_{s}$ items
$\mathbb{M}_{s,I} \in \mathbb{R}^{C_{s} \times D}$ for the source domain data, the visible feature center memory with $C_{t,V}$ items $\mathbb{M}_{t,V} \in \mathbb{R}^{C_{t,V} \times D}$ and the infrared feature center memory with $C_{t,I}$ items $\mathbb{M}_{t,I} \in \mathbb{R}^{C_{t,I} \times D}$ for the target domain data, respectively.
\subsection{Domian-Shared Learning Strategy in Pre-training stage}
The significant inter-domain modality differences between data across the source and target domains usually result in a fact that the VI-ReID network, pre-trained on the source domain data, usually obtains undesirable results when directly applied to the target domain data. Considering that, a novel  Domain-Shared Learning Strategy (DSLS) is designed in this subsection to address such an issue by exploring domain-shared information between the source and target domains.

Specifically, DSLS aims at exploring the domain-shared information to bridge the source and target domains in the pre-training stage. For that, DSLS first introduces target domain data $\hat{D}_{t}$ along with source domain data $D_{s}$ for assisting the pre-training of the VI-ReID network. Then, DSLS extracts the features from source domain data $D_{s}$ and target domain data $\hat{D}_{t}$ through two identical VI-ReID networks $f_{\phi}(*, \phi)$ , \emph{i.e.}, 
\begin{equation}
\begin{split}
&\mathbf{F}_{s, V} = f_{\phi}(x_{s, V}, \phi), \mathbf{F}_{s, I} = f_{\phi}(x_{s, I}, \phi), \\  & 
 \mathbf{F}_{t, V} = f_{\phi}(x_{t, V}, \phi), \mathbf{F}_{t, I} = f_{\phi}(x_{t, I}, \phi), 
\end{split}
\end{equation}
where $\mathbf{F}_{s, V} \in \mathbb{R}^{N_{s,V} \times D}$,  $\mathbf{F}_{s, I} \in \mathbb{R}^{N_{s, I} \times D}$, $\mathbf{F}_{t, V} \in \mathbb{R}^{N_{t, V} \times D}$ and $\mathbf{F}_{t, I} \in \mathbb{R}^{N_{t,I} \times D}$ denote the person features of visible images and infrared images within the source and target domains, respectively. 

After that, a novel Domain-Shared Adversarial Loss (DSAL) will be specially designed to constrain the identity distances across the two domains in an adversarial learning way \cite{GAN}, which can diminish the disparities between
the features from the source domain and those from the target
domain. As well, DSAL will also preserve the distinctions among different
person identities in both the source and target domains for
facilitating the extraction of domain-shared information, \emph{i.e.},
\begin{equation}
\mathcal{L}_{DSAL}=\operatorname{DSAL}(\mathbf{F}_{s, V},  \mathbf{F}_{s, I}, \mathbf{F}_{t, V}, \mathbf{F}_{t, I}, y_{s, V}, y_{s, I}, {y}_{t, V}, {y}_{t, I}).
\end{equation}
It should be noted that the labels $y_{t, V}$ and  $y_{t, I}$ for target domain data are absent. Therefore, a Cluster Refinement with Multiple Results (CRMR) module will be further designed to generate their pseudo labels $\hat{y}_{t, V}$ and  $\hat{y}_{t, I}$  for target domain data to replace ${y}_{t, V}$ and ${y}_{t, I}$ before each training epoch, which will be described in detail subsequently. 

\subsubsection{Domain-Shared Adversarial Loss (DSAL)}
%DSAL enhances the capability of the VI-ReID network in extracting domain-shared features between the source and target domain data in an adversarial learning way \cite{GAN}. 
%Adversarial learning aims at training a generator and a discriminator in a competitive process to achieve a state of equilibrium between the two components. 
For adversarial learning, 
DSAL first treats the VI-ReID network $f_{\phi}( \ast, \phi)$ as the generator, and then designs a classifier ${Class}(\ast, \varrho)$ with its parameters $\varrho$ as the discriminator. As shown in Figure \ref{fig_agw} (b), the discriminator takes the features extracted from the input images by the VI-ReID network as inputs and outputs the probability of the inputs belonging to the target domain. During the training processes of discriminator ${Class}(\ast, \varrho)$, we fix the parameters of the generator $f_{\phi}( \ast, \phi)$ and optimize the discriminator ${Class}(\ast, \varrho)$ by using the cross-entropy loss, \emph{i.e., }
\begin{equation}
\begin{split}
\label{ld}
L_{d}=-\frac{1}{N_{all}} \sum_{n=1}^{N_{all}}{(dom_n\mathrm{log}(Class(f_{\phi}( x^{n}, \phi), \varrho))} \\
+{(1-dom_n)\mathrm{log}(1-Class(f_{\phi}( x^{n}, \phi), \varrho)))} ,
\end{split}
\end{equation}
where $dom_{n}$ denotes the domain label of the sample $x^{n}$. $N_{all}=N_{s, V}+N_{s, I}+N_{t, V}+N_{t, I}$ is the total number of samples in the source and target domains. When training the generator $f_{\phi}( \ast, \phi)$, we fix the parameters of discriminators ${Class}(\ast, \varrho)$ and optimize the generator $f_{\phi}( \ast, \phi)$ to prevent the discriminator from distinguishing the right domains, \emph{i.e.}, 
\begin{equation}
\label{lg}
L_{g}=\frac{1}{N_{all}} \sum_{n=1}^{N_{all}}{(Class(f_{\phi}( x^{n}, \phi), \varrho)-0.5)^{2}}.
\end{equation}

Besides, some contrastive losses are also performed on the generator to preserve the distinctions among different
person identities. Different from conventional contrastive losses that separately constrain features in the source and target domains, our DSAL \textbf{constraints all visible and infrared feature distances within a unified feature space}, thereby explicitly promoting the extraction of domain-invariant features. Specifically, DSAL first concatenates the visible (infrared) feature center memories  $\mathbb{M}_{s, V}$ and $\mathbb{M}_{t, V}$ (or $\mathbb{M}_{s,I}$ and $\mathbb{M}_{t, I}$) from the source domain data and target domain data, obtaining:
\begin{equation}
\label{unitspace}
\begin{split}
& \mathbb{M}_{st,V} = \operatorname{Cat} (\mathbb{M}_{s,V}, \mathbb{M}_{t,V}), \\ &
\mathbb{M}_{st,I} = \operatorname{Cat} (\mathbb{M}_{s,I}, \mathbb{M}_{t,I}),
\end{split}
\end{equation}
where $\mathbb{M}_{st, V}$ and $\mathbb{M}_{st, I} $ denote the concatenated visible and infrared feature center memories, respectively. $ \operatorname{Cat}(*)$ denotes the concatenation operation. 

Then, DSAL introduces the intra-modality constraints on the visible features of the source domain data in the domain-shared feature space $\mathbb{M}_{st,V}$, 
%It constrains the visible features within the source domain data to be closer to their corresponding visible feature centers of the same identifies and far away from the visible feature centers of other identities,
\emph{i.e.},
\begin{equation}
\mathcal{L}_{g}^{s-V}=-\mathrm{log}\sum_{n=1}^{N_{s,V}} \frac{e^{( \mathbf{F}_{s, V}^{n}   \mathbb{M}_{st,V}^{T}[y_{s,V}^{n}] )}}{\sum_{k=1}^{C_{s}+C_{t,V}} e^{(\mathbf{F}_{s, V}^{n}   \mathbb{M}_{st,V}^{T}[k]) }},
\end{equation}
where $\mathbf{F}_{s, V}^{n}$ represents the features of the $n$-th visible image $x_{s, V}^{n}$ in the source domain, $(*)^{T}$ means the transpose operation. Similarly, a constraint $\mathcal{L}_{g}^{s-I}$ is also performed on the infrared features within the source domain datain the domain-shared space $\mathbb{M}_{st,I}$, \emph{i.e.},
\begin{equation}
	\mathcal{L}_{g}^{s-I}=-\mathrm{log}\sum_{n=1}^{N_{s,I}} \frac{e^{( \mathbf{F}_{s, I}^{n} \mathbb{M}_{st,I}^{T}[ y_{s,I}^{n}] )}}{\sum_{k=1}^{C_{s}+C_{t,I}} e^{(\mathbf{F}_{s, I}^{n}   \mathbb{M}^{T}_{st,I}[k]) }}.
\end{equation}

After that, DSAL also builds the inter-modality constraints for the source domain data in the domain-shared feature space to better utilize information in the source domain 
%.
%Specifically, it constrains the visible (infrared) features within the source domain data to be closer to their corresponding infrared (visible) feature centers of the same identities and far away from the infrared (visible) centers of other identities,
 \emph{i.e.},
\begin{equation}
 \mathcal{L}_{g}^{s-VI}=-\mathrm{log}\sum_{n=1}^{N_{s,V}} \frac{e^{( \mathbf{F}_{s, V}^{n}   \mathbb{M}^{T}_{st,I}[ y_{s, V}^{n}] )}}{\sum_{k=1}^{C_{s}+C_{t,I}} e^{( \mathbf{F}_{s, V}^{n_{s,V}} \mathbb{M}^{T}_{st,I}[ k]) }} \\  
-\mathrm{log}\sum_{n=1}^{N_{s,I}} \frac{e^{( \mathbf{F}_{s, I}^{n}     \mathbb{M}^{T}_{st,V}[ y_{s, I}^{n}] )}}{\sum_{k=1}^{C_{s}+C_{t,V}} e^{( \mathbf{F}_{s, I}^{n}  \mathbb{M}^{T}_{st,V}[k]) }}.
\end{equation}

What's more, DSAL also performs some intra-modality constraints on the target domain data, \emph{i.e.},
\begin{equation}
\mathcal{L}_{g}^{t-V}=-\mathrm{log}\sum_{n=1}^{N_{t,V}} \frac{e^{( \mathbf{F}_{t, V}^{n}   \mathbb{M}^{T}_{st,V}[ \hat{y}_{t,V}^{n}] )}}{\sum_{k=1}^{C_{s}+C_{t,V}} e^{(\mathbf{F}_{t, V}^{n}   \mathbb{M}^{T}_{st,V}[ k]) }},
\end{equation}
\begin{equation}
	\mathcal{L}_{g}^{t-I}=-\mathrm{log}\sum_{n=1}^{N_{t,I}} \frac{e^{( \mathbf{F}_{t, I}^{n}   \mathbb{M}^{T}_{st,I}[ \hat{y}_{t,I}^{n}] )}}{\sum_{k=1}^{C_{s}+C_{t,I}} e^{(\mathbf{F}_{t, I}^{n}   \mathbb{M}^{T}_{st,I}[k]) }}.
\end{equation}
%where $\alpha$  is the weight of inter-modality loss in the source domain. 
It is important to note that in this stage, the absence of ground truth in the target domain will result in significant noise among the visible and infrared modalities. Consequently, we have refrained from introducing inter-modality constraints in the target domain.

Accordingly, DSAL is expressed by 
\begin{equation}
\mathcal{L}_{DSAL} = \mathcal{L}_{d} + \mathcal{L}_{g} + \mathcal{L}_{g}^{s-V} + \mathcal{L}_{g}^{s-I} + \mathcal{L}_{g}^{s-VI} + \mathcal{L}_{g}^{t-V}+\mathcal{L}_{g}^{t-I}.
\end{equation}

In summary, by employing $L_{g}$ and $L_{d}$ with Eq. (\ref{ld}) and Eq.(\ref{lg}), DSAL pulls close the source and target domains in an adversarial learning way. Meanwhile, via the unified feature space designed in Eq. (\ref{unitspace}), DSAL can constrain all pedestrian embeddings in a shared feature space, which further effectively mitigates inter-domain modality discrepancies in the UDA-VI-ReID.

\subsubsection{Cluster Refinement with Multiple Results (CRMR)}
\begin{figure}[t]
\centering
\includegraphics[width=3in]{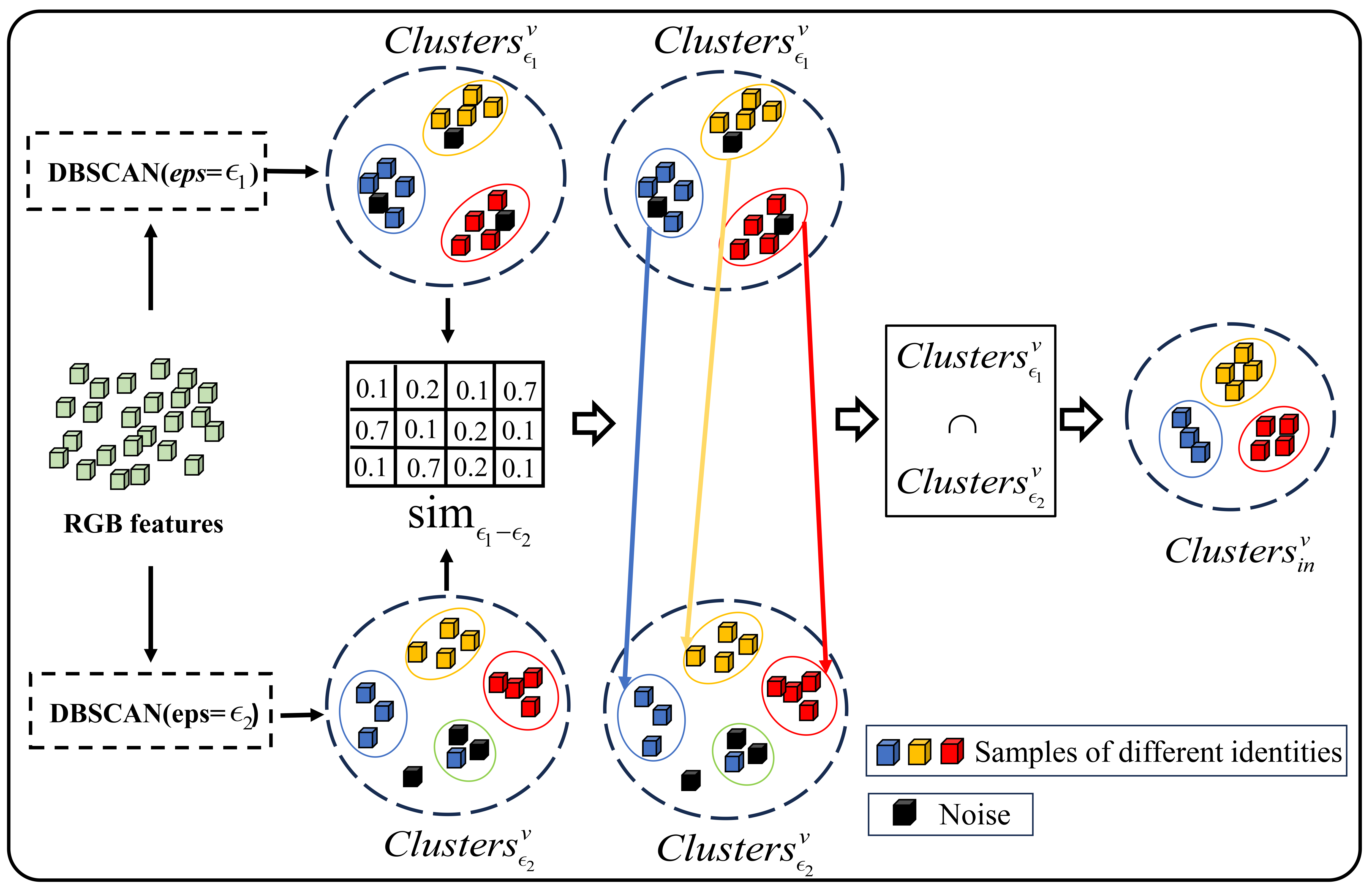}
\caption{
Illustration of the proposed CRMR module. Samples in different colors represent different person identities.
}
\label{fig_cmcr}
\end{figure}

As shown in Fig. \ref{fig_cmcr}, CRMR aims at generating pseudo labels for visible and infrared data, respectively, in the target domain to assist the pre-training of the VI-ReID network. Specifically, we find that DBSCAN\cite{Ester_Kriegel_Sander_Xu_1996}, a widely-used clustering algorithm for pseudo labels generation in unsupervised ReID\cite{Yixiao_Chen_Li_2020, He_Shen_Guo_Ding_Guo_2022}, is very sensitive to the hyperparameter neighborhood radius $esp$, \emph{i.e.}, the maximum distance for one sample to be considered as in the neighborhood of the other sample. Generally speaking, a smaller $esp$ can achieve better intra-cluster consistency, but it is easier to separate the samples of the same identity into different clusters. A larger $esp$ usually results in fewer cluster centers and may cluster the samples of different identities into the same identity. Moreover, the clusters obtained from a small $esp$ and those from a large $esp$ are complementary to each other to some extent. Based on that, our proposed CRMR will explore the mutual relations among multiple clustering results obtained by using different $esp$ settings to yield more reliable pseudo labels. Considering the way of generating pseudo labels for visible data is similar to that for infrared data, we take the visible data as an example for better explanation in the following sections. 

Specifically, given the visible images in the target domain $D_{t, V}$, our proposed CRMR first extracts their  features $\mathbf{F}_{t, V} \in \mathbb{R}^{N_{t,V}\times D}$ by 
\begin{equation}
	\begin{split}
		\mathbf{F}_{t, V} = f_{\phi}(x_{t, V}, \phi).
	\end{split}
\end{equation} 

Then, two sets of clustering results ($Clusters_{\epsilon_{1}}$ and $Clusters_{\epsilon_{2}}$) will be obtained by feeding the visible features  $\mathbf{F}_{t,V}$ into the DBSCAN algorithm with two different $esp$ values, \emph{i.e.}, $\epsilon_{1}$ and $\epsilon_{2}$ (suppose that $\epsilon_{1}$ $>$ $\epsilon_{2}$), respectively, \emph{i.e.},
\begin{equation}
\begin{split}
&Clusters^{v}_{\epsilon_{1}} = \mathrm{DBSCAN}(\mathbf{F}_{t,V}, \epsilon_{1}), \\ & Clusters^{v}_{\epsilon_{2}} = \mathrm{DBSCAN}( \mathbf{F}_{t,V}, \epsilon_{2}). 
\end{split}
\end{equation} 
Here, $Clusters_{\epsilon_1}^{v}$ and $Clusters_{\epsilon_2}^{v}$ contain $K_{\epsilon_{1}}$ and $K_{\epsilon_{2}}$ clusters, respectively and $K_{\epsilon_{1}}$ $\le$ $K_{\epsilon_{2}}$. 
After that, for each cluster(\emph{e.g.}, the $k$-th cluster) $cluster^{v}_{{\epsilon_{1},k}}$ within $Clusters^{v}_{\epsilon_{1}}$, its centroid $u^{v}_{\epsilon_{1},k} \in \mathbb{R}^{1\times D}$ can be calculated by 
\begin{equation}
\label{fn5}
u^{v}_{\epsilon_{1},k} =\frac{1}{ \mathcal{N}(cluster^{v}_{\epsilon_{1}, k}) } \sum_{\mathbf{F}_{t, V}^{n} \in cluster^{v}_{\epsilon_{1}, k} } {\mathbf{F}_{t, V}^{n}},
\end{equation}
where $\mathcal{N}(*)$ denotes the function that returns the number of the items within the input cluster. Accordingly, the centroids $u^{v}_{\epsilon_{1}} \in \mathbb{R}^{K_{\epsilon_{1}}\times D}$ of all clusters can be obtained by
\begin{equation}
u^{v}_{\epsilon_{1}} = \operatorname{Cat}(u^{v}_{\epsilon_{1},1}, u^{v}_{\epsilon_{1},2}, \cdots, u^{v}_{\epsilon_{1},K_{\epsilon_{1}}}).
\end{equation}
Furthermore, the centroids $u^{v}_{\epsilon_{2}} \in \mathbb{R}^{K_{\epsilon_{2}}\times D}$ of all clusters within the clusters $Clusters_{\epsilon_{2}}$ can also be computed in the same way.

Next, CRMR finds the nearest centroid pairs across the two clusters $Clusters_{\epsilon_{1}}$ and $Clusters_{\epsilon_{2}}$. For the $k$-th centroid in $Clusters_{\epsilon_{1}}$, \emph{i.e.}, $u^{v}_{\epsilon_{1},k}$, its similarities with the $q$-th  centroid in $Clusters_{\epsilon_{2}}$, \emph{i.e.}, $u^{v}_{\epsilon_{2},q}$, will be computed by
\begin{equation}
	\mathrm{sim}_{\epsilon_{1}-\epsilon_{2}}(k,q)=\frac{u^{v}_{\epsilon_{1},k} {u^{v T}_{\epsilon_{2},q}} }{ | u^{v}_{\epsilon_{1},k}  | \times | u^{v}_{\epsilon_{2},q}  | }, 
\end{equation} 
where $\mid \ast \mid $ denotes the magnitude of a vector and $\times$ means numerical multiplication.
A higher similarity between two centroids means that the two centroids belong to the same person identity more probably. Based on that, the nearest centroid index $q^{\ast }$ for $u^{v}_{\epsilon_{1},k}$ will be obtained by 
\begin{equation}
q^{\ast }= \underset{q}{\rm argmax} \ \mathrm{sim}_{\epsilon_{1}-\epsilon_{2}}(k,q). 
\end{equation}
%Accordingly, the nearest centroid pairs of all clusters $Nearest_{\epsilon_{1}-\epsilon_{2}}$ within the clusters $Clusters_{\epsilon_{1}}$ that correspond to the clusters $Clusters_{\epsilon_{2}}$  can be obtained, i.e. $Nearest_{\epsilon_{1}-\epsilon_{2}} = \{Nearest_{\epsilon_{1}-\epsilon_{2}}^k \mid k=1,2, \cdots , K_{\epsilon_{1}}\}$.

%Moreover, the nearest centroid pairs of all clusters $Nearest_{\epsilon_{2}-\epsilon_{1}}$ within the clusters $Clusters_{\epsilon_{2}}$ that correspond to the clusters $Clusters_{\epsilon_{1}}$  can be obtained in the same way.
Following that, the intersection set $cluster_{in,k}$ of $cluster^{v}_{\epsilon_{1},k}$ and $cluster^{v}_{\epsilon_{2}, q^{\ast }}$ is considered as the reliable samples and will be saved, \emph{i.e.},
\begin{equation}
cluster_{in,k}^{v} = cluster^{v}_{\epsilon_{1},k} \cap cluster^{v}_{\epsilon_{2},q^{\ast }}.  
\end{equation}

Finally, we can obtain more reliable clustering results within visible data in the target domain, \emph{i.e.}, $Clusters_{in}^{v}=\{ cluster_{in,k}^{v} \mid k=1,2,\dots, K_{\epsilon_{1}} \}$ and assign the same pseudo labels to those samples in $cluster_{in,k}^{v}$.

\subsection{Gradual Alignment Strategy in Fine-tuning stage} %\uppercase\expandafter{\romannumeral2}}
After obtaining the well-initialized VI-ReID network by using the proposed DSLS, the next step is to fine-tune the VI-ReID network in the target domain to improve its performance. The core of this step is how to find the potential corresponding cross-modality clusters and assign them with the same identity.
However, the large intra-domain modality discrepancies across those unlabeled visible and infrared images will easily result in cross-modality alignment challenges, thus leading to suboptimal results. Considering that, a novel Gradual Alignment Strategy (GAS) will be presented to address such an issue in this stage via a cluster-to-holistic alignment way. 

\begin{figure}
\centering
\includegraphics[width=3in]{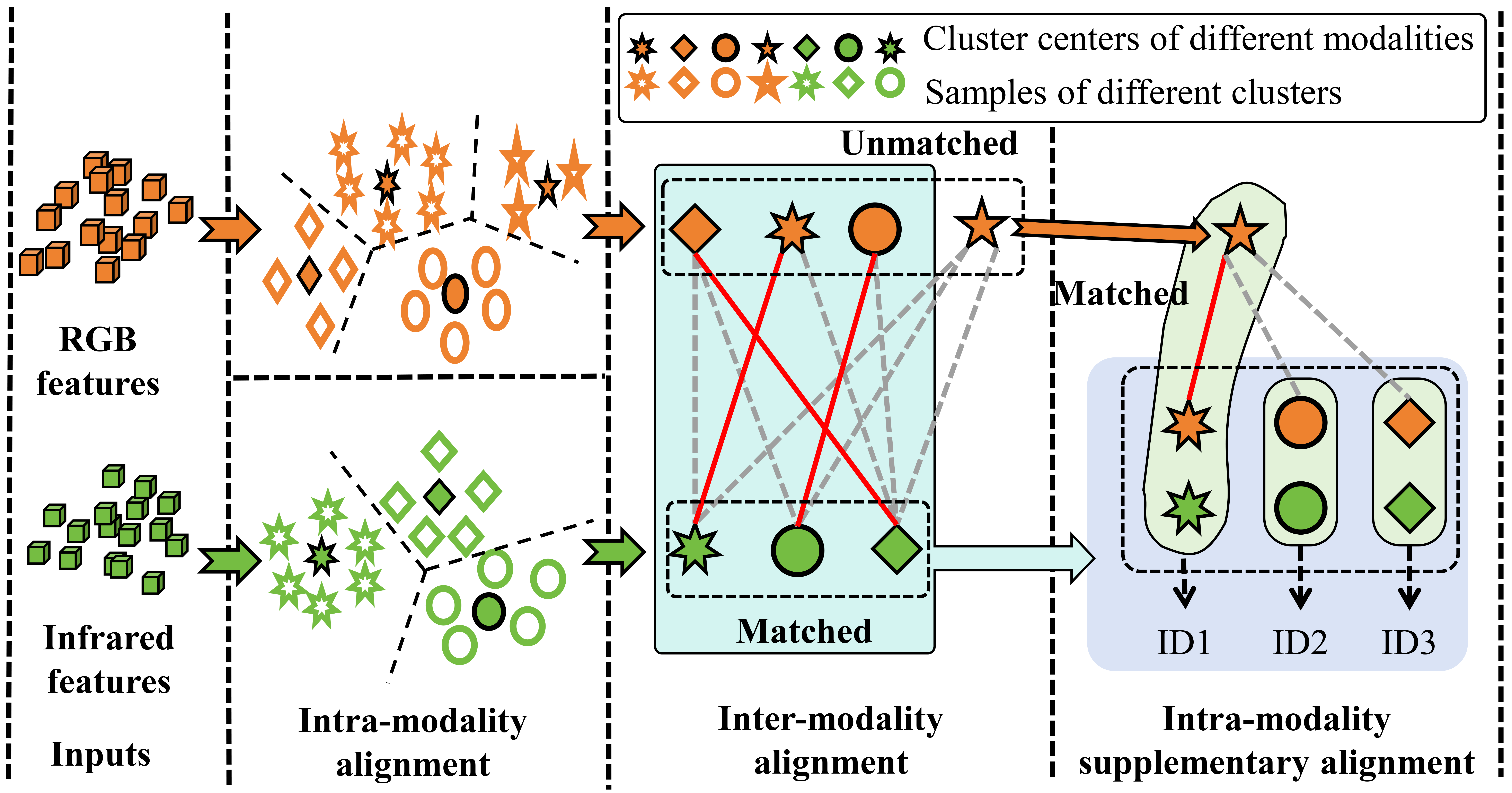}
\caption{
Illustration of the proposed SGM module. It contains three steps: \emph{i.e.}, Intra-modality alignment, Inter-modality alignment and Intra-modality supplementary alignment.
}
\label{fig_SGM}
\end{figure}

As shown in Fig. \ref{fig_SGM}, GAS will first propose a Supplementary Graph Matching (SGM) module to explore the relations among visible and infrared clusters for cluster-level alignment. Accordingly, the pseudo labels for the target domain data can be obtained by assigning those matched results to the same identities, \emph{i.e.}, those matched visible-infrared cluster pairs are expected to belong to the same identities. Then, a Cross-Modality Consistency Constraining (CMCC) module will be further designed to reduce such noise from the misalignments in SGM and, meanwhile, mitigate the intra-domain modality discrepancies in the training process by exploring the holistic interactions among those matched visible-infrared cluster pairs.

\subsubsection{Supplementary Graph Matching (SGM)}
SGM aims to match the visible clusters with the infrared clusters via the following steps.

(1) Intra-modality alignment: Given the target domain data $D_{t} =\left \{ D_{t, V}, D_{t, I}  \right \}$, SGM first clusters the visible samples $D_{t, V} $ and the infrared samples  $D_{t, I}$, respectively, by using CRMR proposed in the pre-training stage. Accordingly, the visible clusters $Clusters_{t}^V$ with $C_{t,V}$ items and infrared clusters $Clusters_{t}^I$ with $C_{t,I}$ items are obtained by
\begin{equation}
\begin{split}
Clusters_t^V &=\mathrm{CRMR}(D_{t, V}),\\
Clusters_t^I &=\mathrm{CRMR}(D_{t, I}).
\end{split}
\end{equation}
Accordingly, their center memories $\mathbb{M}_{t,V}$ and $\mathbb{M}_{t,I}$ can also be initialized by using Eq. (\ref{f5}). 

(2)Inter-modality alignment: Considering that the clusters $Clusters_{t}^V$ and  $Clusters_{t}^I$ are unmatched, \emph{e.g.}, the first cluster from $Clusters_{t}^V$ and that from $Clusters_{t}^I$ usually belong to different person identities, SGM employs the Hungarian matching algorithm \cite{hungarian} to find the optimal matching result
\begin{equation}
\textbf{m}=\left \{ \textbf{m}({n_{t,V}},{n_{t,I})\mid n_{t,V}=1,2,..., C_{t,V};n_{t,I}=1,2,...,C_{t,I}}\right \} \in \mathbb{R}^{1\times{C_{t,V} C_{t, I}}}
\end{equation}
for the clusters $Clusters_{t}^V$ and $Clusters_{t}^I$. Here, $\textbf{m}(n_{t,V}, n_{t,I})$ indicates whether the $n_{t,V}$-th cluster within $Clusters_{t}^V$ and the $n_{t,I}$-th cluster within  $Clusters_{t}^I$ belong to the same person identity ($\textbf{m}(n_{t,V}, n_{t,I})=1)$ or not ($\textbf{m}(n_{t,V}, n_{t,I})=0$).
%
%For that, SGM first conducts the visible graph $\mathcal{G}_V$ with $C_{t, V}$ nodes by using  $\mathbb{M}_{t,V}$ and the infrared graph $\mathcal{G}_I$ with $C_{t,I}$ nodes by using  $\mathbb{M}_{t,I}$. Here, the nodes of  the visible graph $\mathcal{G}_V$ and those of the infrared graph $\mathcal{G}_I$ are denoted by $ [\mathcal{V} ] =  \{u_V^{n_{t,V}}|n_{t,V} = 1,2,...,C_{t,V} \} $ 
%and $[\mathcal{I}] =\{u_I^{n_{t,I}}|n_{t,I} = 1,2,...,C_{t, I}\}$, respectively, where $u_v^{n_{t,V}}=\mathbb{M}_{t,V}[n_{t,V}]$  and $u_I^{n_{t,I}}=\mathbb{M}_{t,I}[n_{t,I}]$. Our target is to find the nearest
%correspondence in $\mathcal{G}_V(\mathcal{G}_I)$ for each node in $\mathcal{G}_I(\mathcal{G}_V)$.
% 
%Then, SGM computes the assignment cost matrix $\mathbb{O}=\left \{ \textbf{O}({n_{t,V}},{n_{t,I})\mid n_{t,V}=1,2,..., C_{t,V};n_{t,I}=1,2,...,C_{t,I}}\right \} \in \mathbb{R}^{1\times{C_{t,V} C_{t, I}}}$ for the Hungarian matching algorithm.
%Here, $\textbf{O}\left(n_{t,V}, n_{t,I}\right)$ represents the distance between the nodes $u_V^{n_{t,V}}$ and $u_I^{n_{t,I}}$, and is computed by
%\begin{equation}
%\textbf{O}\left(n_{t,V}, n_{t,I}\right)= -{\mathrm{exp}}({\frac{u^{n_{t,V}}_{V}  {u^{n_{t,I}}_{I}}^{T} }{ | u^{n_{t,V}}_{V}   |\times  | u^{n_{t,I}}_{I}  | }}).
%\end{equation}

Accordingly,
through the optimization of the Hungarian matching algorithm, SGM can obtain an optimal result $\textbf{m}  $, which can find matched visible-infrared cluster pairs belonging to the same person identities at the lowest cost. Accordingly, those matched pairs will be assigned to the same pseudo labels. 

(3) Intra-modality supplementary alignment: The intra-domain modality discrepancies between visible and infrared modalities of the target domain often lead to different numbers of clusters within the two modalities as discussed in \cite{wu2023unsupervised}, \emph{i.e.}, $C_{t, V} \ne C_{t, I}$. Taking $C_{t, V}>C_{t, I}$ as an example, all the infrared clusters will find their matched visible clusters after the last inter-modality alignment step, while the rest of the visible clusters remain unlabeled. Considering that, SGM further designs an intra-modality supplementary alignment step to explore those unmatched visible clusters for supplementing inter-modality alignment. The basic idea is to re-assign such unmatched visible clusters to those matched visible clusters, since those unmatched visible clusters may be over-clustered and have higher probabilities of belonging to the same identity as their nearest visible clusters. 

Specifically, we suppose that the visible clusters $Clusters_{t}^V$ are divided into matched clusters $Clusters_{t, M}^V$, and unmatch clusters $Clusters_{t, Um}^V$ with $C_{t, V, M}$ items and $C_{t, V, Um}$ items, respectively. Here, $C_{t,V, M} + C_{t,V, Um} = C_{t,V}$. Their corresponding center memories $\mathbb{M}_{t, V}^{M}$ and $\mathbb{M}_{t, V}^{Um}$ can also be obtained in a similar way as in Eq. (\ref{f5}).
For those unmatched clusters, we first build the assignment cost matrix $\mathcal{K}=\{ \textbf{K}(n_{t,V, Um}, n_{t,V, M} )\mid n_{t,V, Um} = 1,2,..., C_{t,V, Um}; n_{t,V, M} = 1,2,..., C_{t,V, M} \} \in \mathbb{R}^{1\times{C_{t,V,Um} C_{t, V,M}}} $.
Here, $\textbf{K}(n_{t,V, Um}, n_{t,V, M})$ denotes the distance between $\mathbb{M}_{t,V}^{Um}[n_{t,V, Um}]$ and $\mathbb{M}_{t,V}^{M}[n_{t,V, M}]$, which is computed by
\begin{equation}
\begin{split}
&\textbf{K}(n_{t,V, Um},n_{t,V, M}) \\
&= \beta (1-\mathbb{M}_{t,V}^{Um}[n_{t,V, Um}]   \mathbb{M}_{t,V}^{M}[n_{t,V, M}]^{T}) \\ & +(1-\beta)d_{J} (\mathbb{M}_{t,V}^{Um}[n_{t,V, Um}], \mathbb{M}_{t,V}^{M}).\end{split}
\label{f29}
\end{equation}
Here, $d_{J}(*)$ denotes the mean Jaccard distance \cite{Ye_Shen_Lin_Xiang_Shao_Hoi_2022} between $\mathbb{M}_{t,V}^{Um}[n_{t,V, Um}]$ and its k-reciprocal nearest neighbors in $\mathbb{M}_{t,V}^{M}$, which provides another distance measurement method to align the intra-domain clusters with high similarity. $\beta$ is the parameter for weighting the two distances and is experimentally set to 0.2 in this paper.

Then, the corresponding nearest matched visible cluster index $n_{t,V, M}^{*}$ for the $n_{t,V, Um}$-th unmatched cluster is found by optimizing
\begin{equation}
\begin{split} 
n_{t,V, M}^{*} = \underset{n_{t,V, M}}{\operatorname{arg \ min}}\  &\textbf{K}(n_{t,V, Um}, n_{t,V, M}). \\
\end{split}
\end{equation}
Finally, the label of the $n_{t,V, Um}$-th unmatched cluster will be assigned by 
\begin{equation}
\hat{y}_{t, V}^{Um}[n_{t,V, Um}]  =  \left\{\begin{matrix}
\hat{y}_{t, V}^{M}[n_{t,V, M}^*], if  \ \textbf{K}(n_{t,V, Um}, n_{t,V, M}^*) < \rho ,
\\
\\
-1, \mathrm{otherwise}.
\end{matrix}\right.
\label{f30}
\end{equation}
Here, $\rho $ denotes the distance threshold. It can be seen that if the distance between $\mathbb{M}_{t,V}^{Um}[n_{t,V, Um}]$ and $\mathbb{M}_{t,V}^{M}[n_{t,V, M}^{*}]$ is less than $\rho $, we believe that its corresponding supplement matched clusters are reliable. Otherwise, the pseudo labels of those unmatched clusters are set to -1, which will be discarded during training.
By doing so, our proposed SGM will explore the relationships among visible and infrared clusters and achieve cluster-level alignment, thus overcoming the intra-domain modality discrepancies to some extent and obtaining corresponding pseudo labels with $C_t$ identities. Here, after filtering out those unreliable pseudo labels, $C_t\le \operatorname{min} (C_{t,V}, C_{t,I})$ represents the number of person identities that simultaneously exist in both the visible and infrared modalities.
Accordingly, the two memeories $\mathbb{M}_{t,V}$ and $\mathbb{M}_{t,I}$ are further updated by Eq. (\ref{fup}) for subsequent steps, thus obtaining $\mathbb{M}_{t,V} \in \mathbb{R}^{C_t \times D}$ and $\mathbb{M}_{t,I} \in \mathbb{R}^{C_t \times D}$.

\subsubsection{Cross-Modality Consistency Constraining (CMCC)}
\begin{figure}
\centering
\includegraphics[width=2.5in]{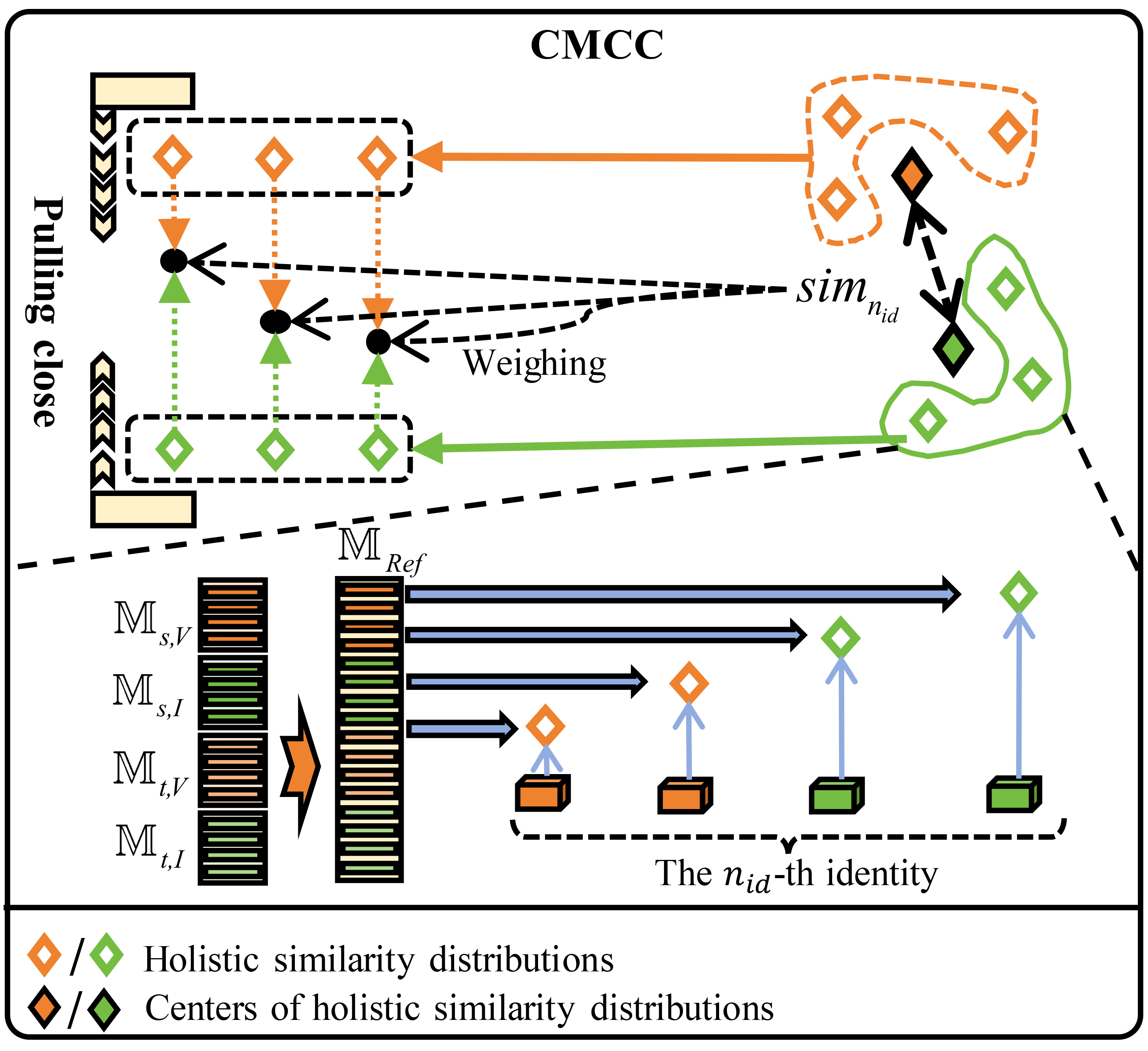}
\caption{
Illustration of the proposed CMCC. Items in different colors represent different person identities.
}
\label{fig_cmcc}
\end{figure}

After obtaining the visible-infrared cluster pairs for the target domain data by using SGM, a Cross-Modality Consistency Constraining (CMCC) module, as shown in Fig. \ref{fig_cmcc}, is further designed to suppress those mismatched
clusters that inevitably exist in the matching results and achieve holistic-level alignment. The basic idea is to use all the person information across visible and infrared modalities of the source and target domain data (termed holistic referring information) to assess the confidence level of each matched visible-infrared cluster pair based on the principle that a smaller distance between a pair of visible and infrared clusters indicates higher reliability.

%the distances between the visible and infrared samples of the same person identity should be smaller than those of different identities and those visual-infrared cluster pairs with smaller distances are regarded more reliable matching results, thus should receive more attention.
%The basic idea is that the center memories $\mathbb{M}_{s, V}$, $\mathbb{M}_{s, I}$, $\mathbb{M}_{t, V}$, and $\mathbb{M}_{t, I}$ altogether contain the global information about every identity in the source and target domain data. Accordingly, we can merge them as a holistic referring space, \emph{i.e.}, $\mathbb{M}_{Ref} \in \mathbb{R}^{C_{Ref} \times D}$ to help assess the confidence level of each visual-infrared cluster pair. Here, $C_{Ref} = C_{s}+C_{s}+C_{t}+C_{t}$. This approach helps identify samples with incorrect pseudo labels by providing a holistic view of all samples from both source and target domains, rather than evaluating them in isolation within a single cluster.

Specifically, CMCC first constructs the holistic referring information $\mathbb{M}_{Ref} \in \mathbb{R}^{C_{Ref} \times D}$ ($C_{Ref} = C_{s}+C_{s}+C_{t}+C_{t}$) by concatenating the center memories $\mathbb{M}_{s, V}$, $\mathbb{M}_{s, I}$, $\mathbb{M}_{t, V}$ and $\mathbb{M}_{t, I}$ of the source domain data and the target domain data, \emph{i.e.},
\begin{equation}
\mathbb{M}_{Ref} = \operatorname{Cat}(\mathbb{M}_{s,V}, \mathbb{M}_{s,I}, \mathbb{M}_{t,V}, \mathbb{M}_{t,I}).
\end{equation}
Then, given all the visible images $x_{t, V}^{n_{id}}$ and all the infrared images  $x_{t, I}^{n_{id}}$ of the $n_{id}$-th identity, CMCC extracts their features $\mathbf{F}_{t, V}^{n_{id}}$ and $\mathbf{F}_{t, I}^{n_{id}}$, respectively, by using the VI-ReID network, \emph{i.e.},
\begin{equation}
\mathbf{F}_{t,V}^{n_{id}} =  f_{\phi}(x_{t,V}^{n_{id}}, \phi), \mathbf{F}_{t,I}^{n_{id}} =  f_{\phi}(x_{t,I}^{n_{id}}, \phi).
\end{equation}
After that, CMCC computes the similarity between those features and the holistic referring information, thus obtaining holistic
similarity distributions $\mathbf{H}_{t,V}^{n_{id}}$ and $\mathbf{H}_{t,I}^{n_{it}}$ by
%instead of simply employing the extracted features $\mathbf{F}_{t,V}^{n_{id}}$ and $\mathbf{F}_{t,I}^{n_{id}}$ for constraining, CMCC obtains their holistic similarity distributions $\mathbf{H}_{t,V}^{n_{id}}$ and $\mathbf{H}_{t,I}^{n_{it}}$ by
\begin{equation}
\begin{split}
& \mathbf{H}_{t,V}^{n_{id}} =  \operatorname{Softmax}(\mathbf{F}_{t,V}^{n_{id}}   \mathbb{M}_{Ref}^{T}),  \\ &
\mathbf{H}_{t,I}^{n_{id}} =  \operatorname{Softmax}(\mathbf{F}_{t,I}^{n_{id}}   \mathbb{M}_{Ref}^{T}),
\end{split}
\end{equation}
where $\operatorname{Softmax}(*)$ is the softmax function.
Compared with directly using $\mathbf{F}_{t,V}^{n_{id}}$ and $\mathbf{F}_{t,I}^{n_{id}}$, the holistic similarity distributions $\mathbf{H}_{t,V}^{n_{id}}$ and $\mathbf{H}_{t,I}^{n_{id}}$ can exploit all the person information within the source and target domains rather than some identity-limited information within certain clusters, thus effectively assessing those generated pseudo labels. 
Naturally, if the visible cluster and infrared cluster are correctly matched (\emph{i.e.}, they have the right pseudo labels), their holistic similarity distributions should be very close; otherwise, they should be far apart. 

Accordingly, in the next step, CMCC assigns higher confidences/weights to those visible-infrared pairs with closer holistic similarity distributions and lower ones to those with far holistic similarity distributions. For that, CMCC computes the similarity between the holistic similarity distributions of visible-infrared pair of the $n_{id}$-th identity, \emph{i.e.}, $sim_{n_{id}}$, to assess their confidence level, by 
\begin{equation}
\label{34}
sim_{n_{id}}= \frac{\mathbf{H}_{t,V, center}^{n_{id}}   {\mathbf{H}_{t,I, center}^{n_{id}T}}}{ | \mathbf{H}_{t,V, center}^{n_{id}}    | \times  | \mathbf{H}_{t,I, center}^{n_{id}} |}. 
\end{equation}
Here, $\mathbf{H}_{t,V, center}^{n_{id}}$, and  $\mathbf{H}_{t,I, center}^{n_{id}}$ are feature centers of $\mathbf{H}_{t,V}^{n_{id}}$ and $\mathbf{H}_{t,I}^{n_{id}}$, respectively, which can be obtained in a similar way by using Eq. (\ref{fn5}).
Based on such confidence value, CMCC pulls close the holistic similarity distributions $\mathbf{H}_{t,V}^{n_{id}}$ and $\mathbf{H}_{t,I}^{n_{id}}$ of the $n_{id}$-th identity to reduce the large intra-domain modality discrepancies by 
\begin{equation}
\mathcal{L}_{cmcc}^{n_{id}}=sim_{n_{id}} \times \mathrm{KL}(\mathrm{log}(\mathbf{H}_{t,V}^{n_{id}}/\tau), \mathrm{log}(\mathbf{H}_{t,I}^{n_{id}} /\tau  )).
\label{f35}
\end{equation}
Here, $\mathrm{KL}(*)$ denotes the Kullback-Leibler Divergence function and $\tau$ is a temperature hyper-parameter. Accordingly, the $\mathcal{L}_{CMCC}$ for all person identities in the target domain are computed by
\begin{equation}
\mathcal{L}_{CMCC}= \frac{1}{C_{t}} \sum_{n_{id}=1}^{C_{t}} L_{cmcc}^{n_{id}}.
\end{equation}
It can be seen that CMCC will prioritize those correctly matched samples while deprioritizing or disregarding those incorrectly matched samples, thus simultaneously reducing the intra-domain modality discrepancies by holistic-level alignment and alleviating the disturbs from those incorrectly labeled samples during the training process.
The overall loss function of our proposed DSLGA during training is
\begin{equation}
L_{final}=L_{DSAL}+\psi L_{CMCC},
\label{f37}
\end{equation}
where $\psi$ is the parameter for balancing $L_{DSLS}$ and $L_{CMCC}$.

By virtue of SGM and CMCC, our proposed GAS conducts cluster-level alignment as well as holistic-level alignment to achieve reliable cross-modality alignment in the target domain. After being fine-tuned with these pseudo labels, the performance of the VI-ReID network pre-trained on the source domain data can be effectively boosted on the target domain data.

\section{Experiments}
\subsection{Datasets and Evaluation Protocols}
We conduct a new CMDA-XD testing method on top of three existing VI-ReID datasets \emph{i.e.}, SYSU-MM01\cite{Wu_Zheng_Yu_Gong_Lai_2017}, RegDB\cite{Nguyen_Hong_Kim_Park_2017} and LLCM\cite{Zhang_2023_CVPR}, to train and test different UDA-VI-ReID models. As shown in Table \ref{t1}, the CMDA-XD testing method consists of six modes, \emph{i.e.}, SYSUtoLLCM, SYSUtoRegDB, LLCMtoSYSU, LLCMtoRegDB, RegDBtoSYSU and RegDBtoLLCM. Details of the proposed CMDA-XD are as follows.

\subsubsection{Source domain data and target domain data of different modes}
The SYSUtoLLCM mode aims to adapt the domain information from the SYSU-MM01 dataset to the LLCM dataset. For that, its training set first takes the training images and labels of the SYSU-MM01 dataset as the source domain data (including 22252 RGB images and 11909 infrared images of 395 identities) and then treats the training images of the LLCM dataset without their labels as the target domain data (including 16946 RGB images and 13975 infrared images of 713 identities). Accordingly, it takes the testing data of the LLCM dataset as its testing set (including 8680 RGB images and 7166 infrared images of 351 identities). 

The SYSUtoRegDB mode, LLCMtoSYSU mode, LLCMtoRegDB mode, RegDBtoSYSU mode and RegDBtoLLCM mode are similar to SYSUtoLLCM mode, and their corresponding training and testing sets are detailed in Table \ref{t1} and Table \ref{t2}.

\begin{table*}[h]
\centering
\caption{Source and target domain data for different modes of our proposed CMDA-XD testing method.}
\resizebox{0.7\linewidth}{!}{
\begin{tabular}{cclllclllclll}
\hline
\multicolumn{1}{c|}{}                      & \multicolumn{4}{c|}{Source domain}          & \multicolumn{8}{c}{Target domain}                                                       \\ \hline
\multicolumn{1}{c|}{Evaluation modes}  & \multicolumn{4}{c|}{training set}           & \multicolumn{4}{c|}{training set}           & \multicolumn{4}{c}{testing set}           \\ \hline
\multicolumn{1}{c|}{SYSUtoLLCM}        & \multicolumn{4}{c|}{SYSU-MM01 training set} & \multicolumn{4}{c|}{LLCM training set}      & \multicolumn{4}{c}{LLCM testing set}      \\ 
\multicolumn{1}{c|}{SYSUtoRegDB}       & \multicolumn{4}{c|}{SYSU-MM01 training set} & \multicolumn{4}{c|}{RegDB training set}     & \multicolumn{4}{c}{RegDB testing set}     \\ 
\multicolumn{1}{c|}{LLCMtoSYSU}        & \multicolumn{4}{c|}{LLCM training set}      & \multicolumn{4}{c|}{SYSU-MM01 training set} & \multicolumn{4}{c}{SYSU-MM01 testing set} \\ 
\multicolumn{1}{c|}{LLCMtoRegDB}        & \multicolumn{4}{c|}{LLCM training set}      & \multicolumn{4}{c|}{RegDB training set} & \multicolumn{4}{c}{RegDB testing set} \\ 
\multicolumn{1}{c|}{RegDBtoSYSU}        & \multicolumn{4}{c|}{RegDB training set}      & \multicolumn{4}{c|}{SYSU-MM01 training set} & \multicolumn{4}{c}{SYSU-MM01 testing set} \\ 
\multicolumn{1}{c|}{RegDBtoLLCM}        & \multicolumn{4}{c|}{RegDB training set}      & \multicolumn{4}{c|}{LLCM training set} & \multicolumn{4}{c}{LLCM testing set} \\ \hline
\end{tabular}
}
\label{t1}
\end{table*}
\subsubsection{Evaluation modes}
All experiments adhere to the established evaluation protocols commonly used in VI-ReID\cite{Wu_Zheng_Yu_Gong_Lai_2017, Ye_Lan_Li_Yuen_2022}. The evaluation metrics encompass Rank-k accuracy, mean Average Precision (mAP) and the mean Inverse Negative Penalty (mINP) metric introduced in\cite{Ye_Shen_Lin_Xiang_Shao_Hoi_2022}.
% \usepackage{multirow}
\begin{table*}[t]
\centering
\caption{Details of SYSU-MM01, RegDB and LLCM datasets, respectively. }
\resizebox{0.6\linewidth}{!}{
\begin{tabular}{ccccccccc}
\hline
\multicolumn{1}{c|}{}                         & \multicolumn{4}{c|}{Training set}                       & \multicolumn{4}{c}{Testing set}                        \\ \hline
\multicolumn{1}{c|}{\multirow{2}{*}{Dataset}} & \multicolumn{2}{c|}{RGB} & \multicolumn{2}{c|}{Infrared} & \multicolumn{2}{c|}{RGB} & \multicolumn{2}{c}{Infrared} \\ \cline{2-9} 
\multicolumn{1}{c|}{}                         & number       & IDs      & number         & IDs         & number       & IDs      & number         & IDs         \\ \cline{0-8}
\multicolumn{1}{c|}{SYSU-MM01}                & 22252        & 395      & 11909          & 395         & 3803         & 96       & 301            & 96          \\ 
\multicolumn{1}{c|}{RegDB}                    & 2060         & 206      & 2060           & 206         & 2060         & 206      & 2060           & 206         \\ 
\multicolumn{1}{c|}{LLCM}                     & 16946        & 713      & 13975          & 713         & 484         & 351      & 7166           & 351         \\ \hline
\end{tabular}
}
\label{t2}
\end{table*}

\subsection{Implementation Details}
\label{implement_details}
The parameters of our VI-ReID network are initialized by using the pre-trained weights from the ImageNet\cite{deng2009imagenet} dataset. 
During training, each training batch consists of half source domain data and half target domain data. The batch size is set to $16\times6$, \emph{i.e.}, a total of 96 images (16 person identities with 6 samples for each identity). All images are resized to $288\times144$. For different data distributions in different modes, the clustering hyperparameters in CMCR are detailed in Table \ref{hyper_cmcr}. The parameter $\beta $ in Eq. (\ref{f29}) is experimentally set to 0.2 to control the relative contributions of different distances for each item in $\mathcal{K}$. The parameter $\tau$ is set to 0.05 in Eq. (\ref{f35}).
The parameter $\psi$ in Eq. (\ref{f37}) is set to 0.5 to control the weight of $L_{DSLS}$ and $L_{CMCC}$. The parameter $\rho $ in Eq. (\ref{f30}) is set to 0.3 to control the threshold of supplementing clusters. We use the Adam optimizer \cite{adam} to train the model with weight decay 5e-4. The initial learning rate is $3.5\times10^{-3}$ and will be decayed by 10 times for every 20 epochs. The network is trained about 80 epochs. It should be noted that $L_{CMCC}$ is just employed in the last 40 epochs.

\begin{table}[]
\centering
\caption{Hyperparameters of CMCR module under different modes.}
\resizebox{0.5\linewidth}{!}{
\begin{tabular}{c|clcl|clcl}
\hline
                 & \multicolumn{4}{c|}{Visible modality}                & \multicolumn{4}{c}{Infrared modality}              \\ \hline
Evaluation modes & \multicolumn{2}{c|}{$\epsilon_{1}$}   & \multicolumn{2}{c|}{$\epsilon_{2}$}    & \multicolumn{2}{c|}{$\epsilon_{1}$}  & \multicolumn{2}{c}{$\epsilon_{2}$}    \\ \hline
SYSUtoLLCM       & \multicolumn{2}{c}{0.6}  & \multicolumn{2}{c|}{0.57}  & \multicolumn{2}{c}{0.6} & \multicolumn{2}{c}{0.57} \\
SYSUtoRegDB      & \multicolumn{2}{c}{0.33}   & \multicolumn{2}{c|}{0.3} & \multicolumn{2}{c}{0.33} & \multicolumn{2}{c}{0.3} \\
LLCMtoSYSU       & \multicolumn{2}{c}{0.66}  & \multicolumn{2}{c|}{0.63} & \multicolumn{2}{c}{0.6} & \multicolumn{2}{c}{0.57} \\
LLCMtoRegDB      & \multicolumn{2}{c}{0.33}   & \multicolumn{2}{c|}{0.3} & \multicolumn{2}{c}{0.33} & \multicolumn{2}{c}{0.3} \\
RegDBtoSYSU      & \multicolumn{2}{c}{0.66}  & \multicolumn{2}{c|}{0.63} & \multicolumn{2}{c}{0.6} & \multicolumn{2}{c}{0.57} \\
RegDBtoLLCM      & \multicolumn{2}{c}{0.6}  & \multicolumn{2}{c|}{0.57}  & \multicolumn{2}{c}{0.6} & \multicolumn{2}{c}{0.57} \\ \hline
\end{tabular}
}
\label{hyper_cmcr}
\end{table}

\subsection{Comparisons with State-of-the-Arts}
In this subsection, we compare our proposed model with several existing SOTA VI-ReID models on the new testing method CMDA-XD. It should be noted that there are few existing UDA-VI-ReID methods except for our proposed model. Therefore, the other four types of ReID models, including fully-supervised VI-ReID, USL-VI-ReID, UDA-ReID and UDA-VI-ReID models are employed for comparisons. 
%Here, the VI-ReID models include X-Modality \cite{Xmodality}, AGW\cite{Ye_Shen_Lin_Xiang_Shao_Hoi_2022}, DFLN-ViT \cite{zhao2022spatial}, DEEN \cite{llcm}, FMCNet \cite{zhang2022fmcnet}, SGIEL \cite{kim2023partmix} and WRIM \cite{wu2024wrim}.
We also report the results of our proposed DSLGA under fully supervised training settings, \emph{i.e.}, DSLGA$^{*}$. 
%To further evaluate the performance of our proposed model, we also employe a transformer-based backbone,as described
%in \cite{Yang_2024_CVPR}, to validate the effectiveness of our modules \emph{i.e.}, DSLGA-trans.
%The USL-VI-ReID models include OTLA \cite{Wang_Zhang_Chen_Zhang_Wang_Qu_Xie}, ADCA \cite{Yang_Chen}, DOTLA \cite{cheng2023unsupervised}, MBCCM \cite{cheng2023efficient} and CMAM \cite{10538304}. For these VI-ReID and USL-ReID models, we directly employ their results on the target domain dataset for comparisons. The UDA-ReID models include SPCL \cite{SPCL}, MMT \cite{Yixiao_Chen_Li_2020} and SECRET \cite{He_Shen_Guo_Ding_Guo_2022}. 
It should be noted that those UDA-ReID models are designed and trained only for single-modality re-identification. Therefore, we retrain them under the proposed testing method for comparisons.
The evaluation results of different methods are shown in Table \ref{final_results}.
\subsubsection{Comparisons with Supervised Methods}
As shown in Table \ref{final_results}, our proposed model DSLGA$^{*}$ achieves competitive results compared with existing SOTA models under fully supervised settings. However, a large performance margin still exists between our proposed DSLGA under UDA-VI-ReID settings and fully supervised SOTA models. Furthermore, we can also find that our proposed DSLGA under UDA-VI-ReID settings achieves competitive and even better results than those fully supervised models presented before the year 2021, \emph{e.g.} AGW, which demonstrates the feasibility of our proposed UDA-VI-ReID method. It also indicates that the task of UDA-VI-ReID still has enormous potential in the future, appealing to a greater number of scholars who will embark on exploring the topic and contributing to the advancement of this field.

%As shown in Table \ref{final_results}, the proposed DSLGA trained under UDA-VI-ReID task settings surpasses the fully supervised AGW\cite{Ye_Shen_Lin_Xiang_Shao_Hoi_2022} by about 8\%, 0.03\% and 20\% for Rank-1 on the SYSU-MM01 dataset, LLCM dataset and RegDB dataset respectively, which demonstrate the feasibility of our proposed UDA-VI-ReID method. For the UDA-VI-ReID task is currently in the initial stage and its theoretical foundation needs further improvements, our method is not yet competitive compared to the latest results. Besides, we further train the DSLGA under supervised settings \emph{i.e.}, DSLGA$^{*}$. Compared with DEEN, FMCNet, and SGIEL, DSLGA$^{*}$ achieves competitive results, which means the UDA-VI-ReID task has enormous potential in the future.

\subsubsection{Comparisons with Unsupervised Methods}
Experimental results in Table \ref{final_results} demonstrate that our proposed DSLGA outperforms existing unsupervised methods under various settings. 
This proves the effectiveness of our proposed method, which can effectively transfer the learned knowledge from the source domain to the target domain. 
%However, the proposed method does not surpass state-of-the-art unsupervised approaches due to UDA-VI-ReID has only recently begun to be systematically investigated, where significant opportunities for future advancement remain.

\subsubsection{Comparisons with UDA-ReID Methods}
It can be seen that existing UDA-ReID methods perform poorly on our proposed testing method. This indicates that, due to the large inter-domain modality discrepancies and intra-domain modality discrepancies, existing UDA-ReID methods usually struggle with the task of UDA-VI-ReID. Meanwhile, our proposed method achieves significant improvements over existing UDA-ReID methods. This may owe to the fact that our proposed DSLS can effectively reduce the inter-domain modality discrepancies by exploring the domain-shared information between the source and target domains, and the proposed GAS can well handle the large intra-domain modality discrepancies through a cluster-to-holistic alignment strategy. 
\begin{table*}[h]
\caption{Performance comparisons of our proposed method and state-of-the-art methods on CMDA-XD testing method.}
\resizebox{\linewidth}{!}{
\begin{tabular}{cc|ccc|ccc|ccc|ccc|ccc|ccc}
\hline
\multicolumn{2}{c|}{\multirow{2}{*}{}}                     & \multicolumn{3}{c|}{LLCMtoSYSU} & \multicolumn{3}{c|}{RegDBtoSYSU} & \multicolumn{3}{c|}{SYSUtoLLCM}       & \multicolumn{3}{c|}{RegDBtoLLCM}      & \multicolumn{3}{c|}{SYSUtoRegDB}       & \multicolumn{3}{c}{LLCmtoRegDB}       \\ \cline{3-20} 
\multicolumn{2}{c|}{}                                      & \multicolumn{3}{c|}{All Search} & \multicolumn{3}{c|}{All Search}  & \multicolumn{3}{c|}{InfaredtoVisible} & \multicolumn{3}{c|}{InfaredtoVisible} & \multicolumn{3}{c|}{VisibletoInfrared} & \multicolumn{3}{c}{VisibletoInfrared} \\ \hline
\multicolumn{2}{c|}{Backbone}                              & r1        & mAP      & mINP     & r1        & mAP       & mINP     & r1         & mAP        & mINP        & r1         & mAP         & mINP       & r1         & mAP         & mINP        & r1         & mAP        & mINP        \\ \hline
\multicolumn{1}{c|}{\multirow{4}{*}{ }}     & AGW 2021 \cite{Ye_Shen_Lin_Xiang_Shao_Hoi_2022} &      47.5     &    47.65      &    35.3      &     -      &      -     &     -     &      46.4      &       54.8     &      -       &      -      &     -        &      -      &       70.05     &     66.37        &     50.19        &    -        &        -    &     -        \\

\multicolumn{1}{c|}{\multirow{4}{*}{ VI-ReID}}    & FMCNet 2022 \cite{zhang2022fmcnet} &     66.34      &    62.51      &      -    &       -    &          - &      -    &     -       &     -       &            - &     -       &        -     &      -      &     89.12       &    84.43         &     -        &      -      &    -        &      -       \\
\multicolumn{1}{c|}{}                             & DEEN 2023 \cite{llcm} &     74.7      &     71.8     &    -      &      -     &          - &       -   &        54.9    &   62.9         &          -   &    -        &     -       &      -      &     91.1       &      85.1        &       -      &     -       &       -     &      -       \\
\multicolumn{1}{c|}{}                             & WRIM 2024\cite{wu2024wrim} &     77.4      &     75.4     &     -     &    -       &          - &    -      &    -        &      -      &            - &      -      &     -        &      -      &      94.5      &            90.5 &       -      &       -     &    -        &    -         \\
\multicolumn{1}{c|}{}                             & CFSR-Net 2025\cite{CHEN2025111131} &     76.77      &     73.96     &     -     &    -       &          - &    -      &    -        &      -      &            - &      -      &     -        &      -      &      94.87      &            92.22 &       -      &       -     &    -        &    -         \\ 
\multicolumn{1}{c|}{}                             & DSLGA$^{*}$(Ours) &     72.16       &     69.96      &    57.59      &           69.10 &     66.89       &    54.07      &    54.07        &     60.82       &       57.68      &      55.10       &       61.53       &      58.25       &            95.32 &      92.25        &      85.69        &       96.11     &      92.61      &       86.00       \\ \hline
\multicolumn{1}{c|}{\multirow{4}{*}{USL-VI-ReID}} & ADCA 2022 \cite{Yang_Chen} &     45.51      &    42.73      &     28.29      &          - &     -      &      -    &       -     &     -       &       -      &        -    &        -     &     -       &      67.02      &     64.05        &      52.67       &      -      &           - &      -       \\
\multicolumn{1}{c|}{}                             & PGM 2023 \cite{wu2023unsupervised} &     57.27      &     51.78      &     34.96      &          - &     -      &      -    &       -     &     -       &       -      &        -    &        -     &     -       &      69.48      &      65.41        &      - &      -      &           - &      -       \\
\multicolumn{1}{c|}{}                             & DOTLA 2023 \cite{cheng2023unsupervised} &     50.36      &    47.36      &     32.4      &     -      &    -       &    -      &           - &      -      &    -         &      -      &     -        &       -     &     85.63       &      76.71       &      61.58       &           - &        -    &      -       \\
\multicolumn{1}{c|}{\multirow{4}{*}{}}                             & MBCCM 2023 \cite{cheng2023efficient} &     53.14      &    48.16      &     32.41      &   -        &     -      &     -     &       -     &      -      &     -        &      -      &         -    &         -   &      83.79    &      77.87       &     65.04        &        -    &      -      &     -        \\ 
\multicolumn{1}{c|}{}                             & IMSL 2024 \cite{pang2024inter} &     57.96      &    53.93      &     -      &     -      &    -       &    -      &           - &      -      &    -         &      -      &     -        &       -     &     70.08       &      66.30       &      -       &           - &        -    &      -       \\  
\multicolumn{1}{c|}{}                             & SCA-RCP 2024 \cite{li2024inter} &     51.41      &    48.52      &      33.56      &     -      &    -       &    -      &           - &      -      &    -         &      -      &     -        &       -     &     85.59       &      79.12       &      -       &           - &        -    &      -       \\
\multicolumn{1}{c|}{}                             & DMSL 2025 \cite{dai2025dual} &     53.89      &    51.58       &      37.93      &     -      &    -       &    -      &          40.71 &      46.82      &    43.07         &      -      &     -        &       -     &     \textbf{89.68}       &      84.32       &     72.79        &           - &        -    &      -       \\ \hline
%\multicolumn{1}{c|}{}                             & RULN 2025 \cite{teng2025relieving} &     61.87      &    58.92      &      45.01      &     -      &    -       &    -      &           - &      -      &    -         &      -      &     -        &       -     &     88.75       &      82.14       &      68.75       &           - &        -    &      -       \\ \hline
\multicolumn{1}{c|}{UDA-ReID}                             & MMT 2020 \cite{Yixiao_Chen_Li_2020}    & 7.06      & 9.16     & 7.22     & 6.34      & 8.23      & 6.76     &     6.64       &     8.63       &            7.85 &      7.81      &     10.40        &   9.35         &      5.10      &      6.24       &      6.39       &     6.97       &      7.20      &    6.69         \\
\multicolumn{1}{c|}{}    & SPCL 2021 \cite{SPCL}   & 11.17      & 13.23     & 9.35     & 12.39      & 14.73      & 11.01     &    14.45        &    18.68        &            16.14 &      13.27      &      18.35       &      16.35      &      11.02      &     11.50        &       11.19      &     10.68       &     12.58       &     12.22        \\
\multicolumn{1}{c|}{}                             & SECRET 2022 \cite{He_Shen_Guo_Ding_Guo_2022}    & 12.30      & 14.92     & 13.03     & 17.42      & 15.04      & 15.72     &    17.54        &     23.35       &            19.22 &        17.10    &      21.86       &      18.66      &      12.62      &     14.18        &    10.85         &     10.39       &     12.28       &    10.21         \\
\multicolumn{1}{c|}{}                             & DMJL 2023 \cite{CHEN2023109369}    & 11.32      & 12.57     & 12.31     & 15.18      & 14.17      & 12.11 &    15.21        &     18.15       &            18.47 &        18.70    &      23.54       &      19.70      &      14.12      &     16.88        &    11.44         &     9.78       &     10.12       &    8.70         \\ \hline

\multicolumn{1}{c|}{}                             & TAA 2023 \cite{Yang_Chen_Ma_Ye}   &   48.77     &   42.43    &   25.37    &       - &   -     &   -    &   -         &  -          &       -      &     -       &    -         &  -          &    63.23        &     56.00        &            38.77 &       -     &         -   &      -       \\ 
\multicolumn{1}{c|}{UDA-VI-ReID}                  & DSLGA(Ours)   & 54.56     & 49.50    & 34.16    & 55.57     & 50.87     & 35.73    &     46.43       &    52.16        &            48.47 &      44.36      &       50.18      &   46.37         &      87.41      &     81.94        &    70.84         &     89.60       &      84.18      &    72.88   \\   \hline   
%\multicolumn{1}{c|}{}                  & DSLGA-trans(Ours)   & 57.64     & 54.51    & 39.92    & \textbf{58.62}     & \textbf{55.20}     & \textbf{41.98}    &     \textbf{49.43}       &    \textbf{54.16}        &            \textbf{50.12} &      48.32      &       53.84      &   48.20         &      86.43      &     80.12        &    68.87         &     \ 89.63     &      \textbf{86.66}      &    \textbf{74.32}         \\ \hline
\end{tabular}
}
\label{final_results}
\end{table*}
\subsubsection{Comparisons with UDA-VI-ReID Methods}
It can be seen that DSLGA outperforms existing UDA-VI-ReID method TAA by a large margin. This indicates that our proposed DSLS and GAS, which are specifically designed to address intra-domain and inter-domain modality discrepancies, respectively, make sense. This also demonstrates that DSLGA can effectively transfer knowledge from the source domain to the target domain.

\subsection{Ablation Study}
\label{ablation_chapter}
In this subsection, we validate the effectiveness of each component in our proposed DSLGA. As shown in Table \ref{ablation}, several variants of our proposed model are evaluated on the LLCMtoSYSU and RegDBtoSYSU modes of the CMDA-XD testing method. Here,
`RI' denotes the model whose parameters are randomly initialized by using Xavier \cite{glorot2010understanding}, without pre-training and fine-tuning.
`PRE-only' denotes that the VI-ReID network is only trained on the source domain data and tested on the target domain data without fine-tuning.
`DSE', `DSAL' and `CRMR' denote the exploration of domain-shared information, the proposed DSAL and CRMR module in DSLS, respectively. 
`GM' denotes that the Hungarian matching algorithm \cite{hungarian} is used for cross-modality matching in the fine-tuning stage. `SGM' of GAS denotes employing our proposed SGM for cross-modality matching. `CMCC' of GAS denotes that our proposed CMCC module is used for holistic-level alignment. 
It should be noted that all the compared models are trained by using the intra- and inter-modality triple loss \cite{Yang_Chen} in the pre-training and fine-tuning stage, respectively, except for employing DSAL in the pre-training stage and CMCC in the fine-tuning stage. Meanwhile, all the compared models employ  DBSCAN for intra-modality clustering, except for the ones using CRMR. 
\begin{table*}[h]
\caption{Ablation study on the effectiveness of each component in DSLGA.}
\centering
\resizebox{0.8\linewidth}{!}{
\begin{tabular}{c|cccccccc|cccccc}
\hline
\multirow{3}{*}{Index} & \multicolumn{5}{c|}{Pre-training}                                                                                                                                            & \multicolumn{3}{c|}{Fine-tuning}                                                 & \multicolumn{6}{c}{Results}                                                  \\ \cline{2-15} 
                       & \multicolumn{1}{c|}{\multirow{2}{*}{RI}} & \multicolumn{1}{c|}{\multirow{2}{*}{Pre-Only}} & \multicolumn{3}{c|}{DSLS}                                                        & \multicolumn{1}{c|}{\multirow{2}{*}{GM}} & \multicolumn{2}{c|}{GAS}              & \multicolumn{3}{c|}{LLCMtoSYSU}            & \multicolumn{3}{c}{RegDBtoSYSU} \\ \cline{4-6} \cline{8-15} 
                       & \multicolumn{1}{c|}{}                    & \multicolumn{1}{c|}{}                          & \multicolumn{1}{c|}{DSE} & \multicolumn{1}{c|}{DSAL} & \multicolumn{1}{c|}{CRMR} & \multicolumn{1}{c|}{}                    & \multicolumn{1}{c|}{SGM} & CMCC       & r1    & mAP   & \multicolumn{1}{c|}{mINP}  & r1        & mAP      & mINP     \\ \hline
1                      & \checkmark                               &                                                &                          &                           &                           &                                          &                          &            & 1.28  & 3.03  & \multicolumn{1}{c|}{1.69}  & 1.28      & 3.03     & 1.69     \\
2                      &                                          & \checkmark                                     &                          &                           &                           &                                          &                          &            & 5.77  & 7.08  & \multicolumn{1}{c|}{2.80}  & 2.50      & 4.50     & 2.00     \\
3                      &                                          &                                                & \checkmark               &                           &                           &                                          &                          &            & 35.06 & 35.15 & \multicolumn{1}{c|}{21.13} & 35.47     & 35.64    & 22.06    \\
4                      &                                          &                                                & \checkmark               & \checkmark                &                           &                                          &                          &            & 37.92 & 36.83 & \multicolumn{1}{c|}{24.23} & 36.52     & 36.05    & 23.06    \\
5                      &                                          &                                                & \checkmark               & \checkmark                & \checkmark                &                                          &                          &            & 39.25 & 37.67 & \multicolumn{1}{c|}{24.30} & 39.02     & 38.86    & 25.97    \\
6                      &                                          &                                                & \checkmark               & \checkmark                & \checkmark                & \checkmark                               &                          &            & 49.05 & 45.58 & \multicolumn{1}{c|}{30.91} & 51.61     & 46.15    & 30.34    \\
7                      &                                          &                                                & \checkmark               & \checkmark                & \checkmark                &                                          & \checkmark               &            & 50.05 & 47.72 & \multicolumn{1}{c|}{33.24} & 53.13     & 48.64    & 33.05    \\
8                      &                                          &                                                & \checkmark               & \checkmark                & \checkmark                &                                          & \checkmark               & \checkmark & 54.56 & 49.50 & \multicolumn{1}{c|}{34.16} & 55.57     & 50.87    & 35.73    \\ \hline
\end{tabular}
}
\label{ablation}
\end{table*}

\begin{figure}[t]
\centering
\includegraphics[width=3in]{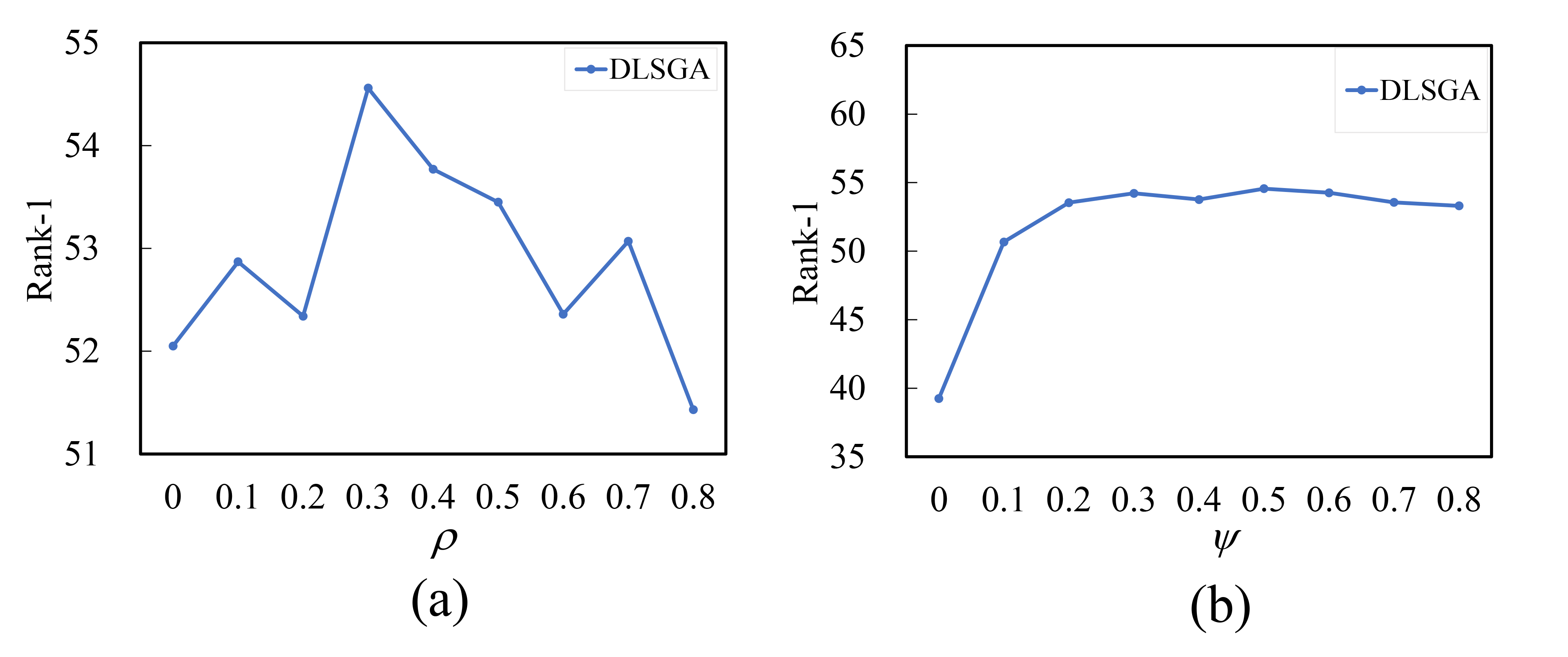}
\centering
\caption{
Evaluation results for the hyperparameters $\rho$ in Eq. (\ref{f30}) and $\psi$ in Eq. (\ref{f37}).
}
\label{PA}
\end{figure}
For the pre-training stage, the results of Index 1 and Index 2 show that the VI-ReID network pre-trained on the source domain dataset is only slightly better than the randomly initialized one. This means that simply pre-training with the source domain data cannot handle the inter-domain modality discrepancies in the UDA-VI-ReID task. 
While the model of Index 3 achieves significant improvements over that of Index 2, which verifies the basic idea of our proposed DSLS, \emph{i.e.}, the exploitation of domain-shared information can effectively reduce the inter-domain modality discrepancies, thus facilitating the transfer of knowledge from the source domain to the target domain. Furthermore, the model of Index 4 further boosts its performance by employing our proposed DSAL. This indicates that DSAL can further help explore more domain-shared information by pulling close the feature distributions between the source and target domain data. Besides, the results of Index 5 also show that our proposed CRMR can achieve better intra-modality clustering and generate more reliable pseudo labels than DBSCAN through its multiple clustering results refinement strategy. Eventually, our proposed DSLS effectively overcomes the inter-domain modality discrepancies, obtaining a well-initialized VI-ReID network for target domain data. 
\begin{figure}
\centering
\includegraphics[width=3in]{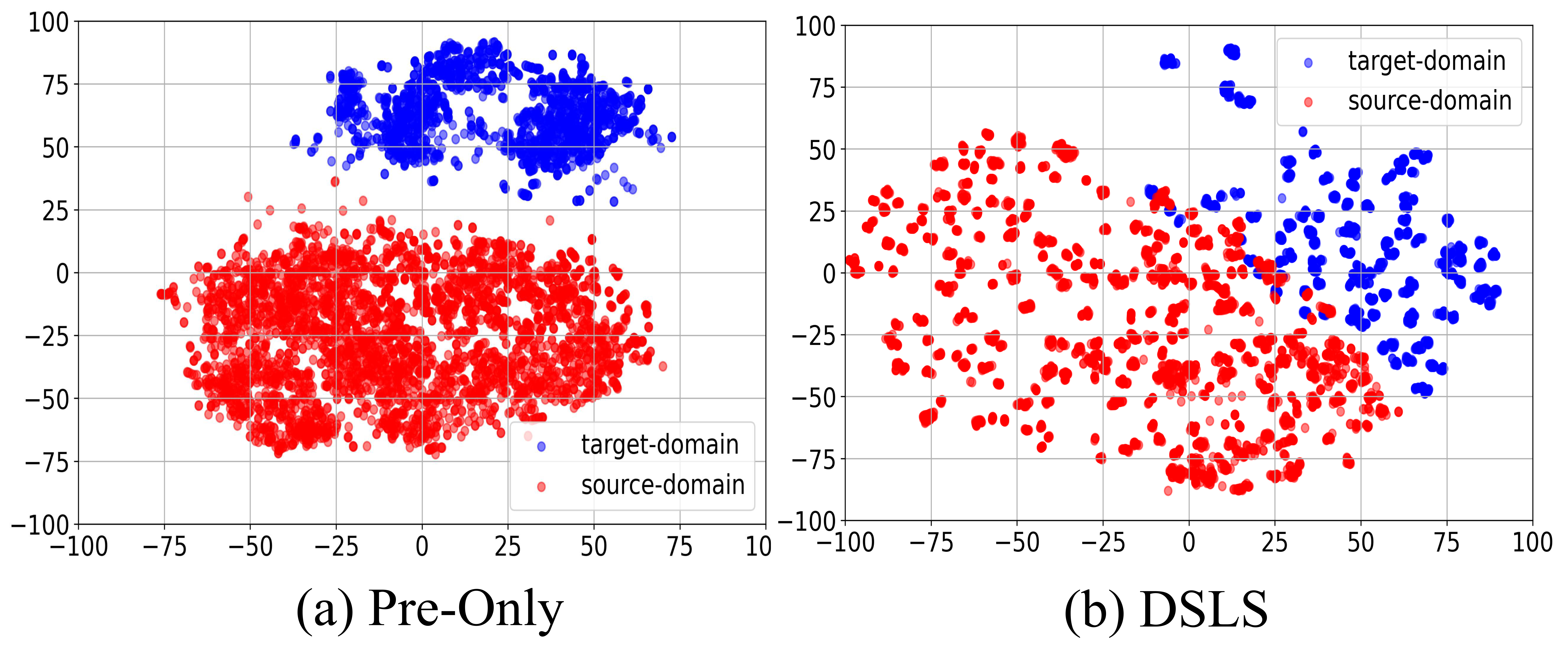}
\centering
\caption{Visualization of inter-domain modality discrepancies. (a) and (b) visualize the distributions of the features before and after employing the DSLS strategy, respectively.
}
\label{vis_inter}
\end{figure}

For the fine-tuning stage, compared with the results of Index 6 and Index 7, it can be seen that our proposed SGM obtains better results than the traditional GM. This means that our proposed SGM can better handle the cross-modality alignment challenges of UDA-VI-ReID than GM does by generating more reliable pseudo labels via its supplementary graph-matching based cluster-level alignment strategy. The model of Index 7, \emph{i.e.}, our final model, achieves the best performance. This indicates that our proposed CMCC can effectively reduce the noisy information caused by mismatched pseudo labels via the holistic-level alignment strategy. Meanwhile, it also indicates that, by virtue of SGM and CMCC, GAS can effectively address the cross-modality alignment challenges caused by the large intra-domain modality discrepancies, thus obtaining better fine-tuning results for the target domain data. 

\subsection{Parameters Analysis}
\subsubsection{Evaluation of $\rho$}
The hyperparameter $\rho$ in Eq. (\ref{f30}) determines the quality of the final cross-modality matching results.
Thus, we evaluate its validity in this subsection on the LLCMtoSYSU dataset under all search evaluation mode. As shown in Fig. \ref{PA} (a), we vary $\rho$ from 0 to 0.8 at intervals of 0.1. The effectiveness of $\rho$ for Rank-1 shows an overall trend of increasing first and then decreasing with the growth of $\rho$. Eventually, it achieves the highest Rank-1 when $\rho=0.3$. Accordingly, we set $\rho=0.3$ in this paper.

\begin{figure}
\centering
\includegraphics[width=3in]{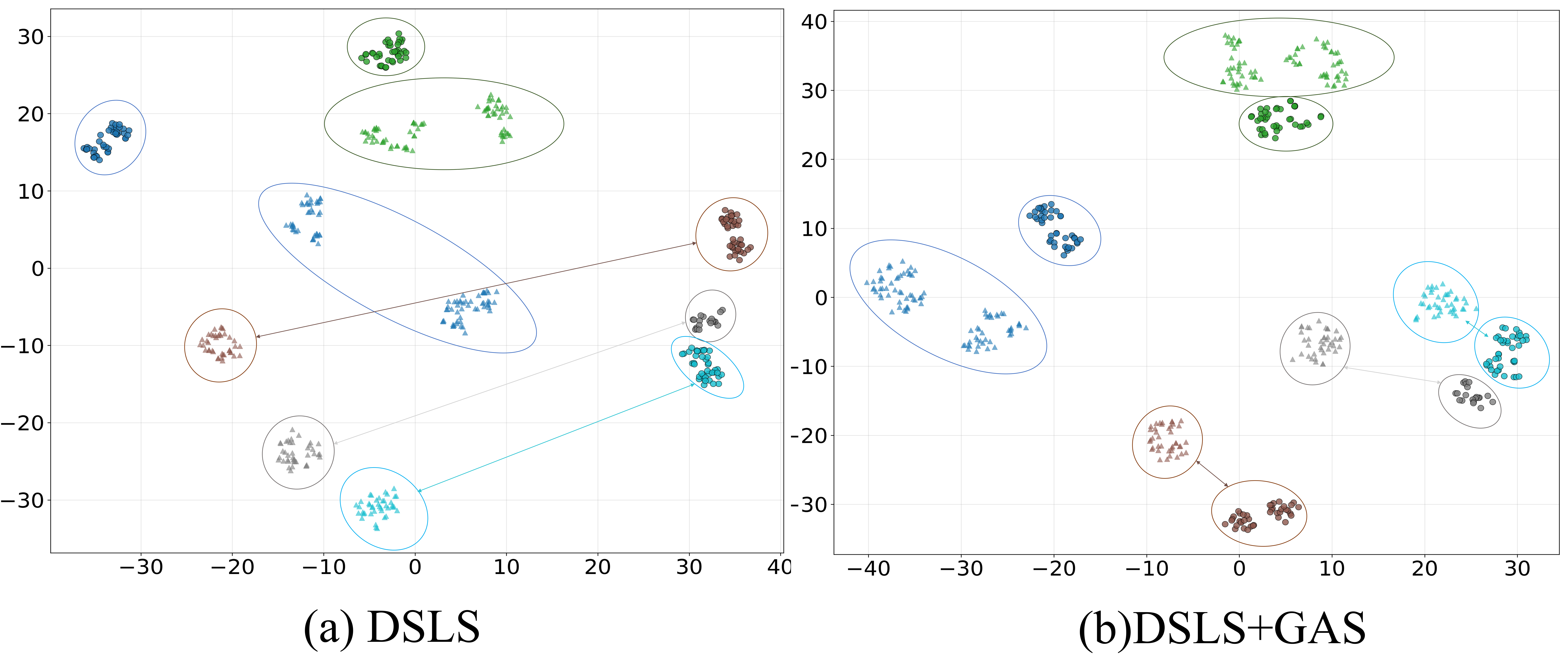}
\centering
\caption{Visualization of intra-domain discrepancies. (a) and (b) visualize the distributions of the features before and after employing the GAS strategy, respectively. Clusters in different colors mean different person identities. The triangular points are samples from the infrared modality, while the circular points represent samples from the visible light modality.}
\label{vis_intra}
\end{figure}

\subsubsection{Evaluation of $\psi$}
The hyperparameter $\psi$ in Eq. (\ref{f37}) influences the optimization process of the VI-ReID network.
Thus, we evaluate its effectiveness on the LLCMtoSYSU mode under all search evaluation mode.
As shown in Fig. \ref {PA} (b), we vary $\psi$ from 0 to 0.8 at intervals of 0.1.
It can be seen that the results of DSLGA for Rank-1 exhibit an overall trend of initially increasing and subsequently stabilizing. Meanwhile, it achieves the highest Rank-1 when $\psi = 0.5$. Therefore, we set $\psi = 0.5$ for our final model in this paper.

\subsubsection{Visualization}
As shown in Fig. \ref{vis_inter} and Fig. \ref{vis_intra}, we visualize the inter-domain and intra-domain discrepancies of different models on the LLCMtoSYSU mode for further verifying the effectiveness of our proposed DLSGA. All experiments are conducted on the testing dataset. As illustrated in Fig. \ref{vis_inter}, the inter-domain distance between the source and target domains under the `DSLS' strategy is smaller than that under the `Pre-Only' strategy. This confirms that the domain-shared features can effectively bridge the source and target domains, thereby reducing the inter-domain modality discrepancies. Meanwhile, as depicted in Fig. \ref{vis_intra}, the feature distances between clusters sharing the same identity but originating from different modalities are reduced after applying our proposed `GAS' strategy. The results indicate that our cluster-to-holistic alignment approach successfully: (1) produces accurate cross-modality pseudo-labels, and (2) filters out mismatched correspondences, consequently reducing intra-domain modality discrepancies.

\section{Conclusion}
This paper presents a novel two-stage model, \emph{i.e.}, Domain-Shared Learning and Gradual Alignment (DSLGA) model, for UDA-VI-ReID, which effectively transfers the knowledge learned from the source domain data to the target domain data without annotating the latter. By virtue of its domain-shared information exploration-based Domain-Shared Learning Strategy (DSLS), DSLGA overcomes the large inter-domain modality discrepancies and pre-trains a well-initialized VI-ReID network for the target domain data by using those labeled source domain data. Owing to the proposed Gradual Alignment Strategy (GAS), DSLGA gradually addresses the cross-modality alignment challenges caused by the intra-domain modality discrepancies in a cluster-to-holistic alignment way, \emph{i.e.}, first generating reliable pseudo-labels in the cluster level by the Supplementary Graph Matching (SGM) module and then suppressing those incorrect pseudo labels in the holistic level by the Cross-Modality Consistence Constraining (CMCC) module. Finally, a new testing method \emph{i.e.}, CMDA-XD, is constructed for training and testing different UDA-VI-ReID models. Extensive experiments demonstrate that our method significantly outperforms existing domain adaptation methods for ReID and even some supervised methods under various settings, pushing UDA-VI-ReID to real-world deployment.

\section{Acknowledgments}
This work is supported by the National Natural Science Foundation of China under Grant No.61803290 and No.61773301, and by the Fundamental Research Funds for the Central Universities under Grant No.ZYTS24022.

\bibliography{ref}

\end{document}